\documentclass[10pt]{article}
\usepackage{amsfonts}
\usepackage{balance}
\usepackage{booktabs}
\usepackage{amsmath}
\usepackage{tabularx}
\usepackage{subfigure}
\usepackage{graphicx}
\usepackage{epstopdf}
\usepackage{multirow}
\usepackage{rotating}
\usepackage{algorithmic}
\usepackage{footnote}
\usepackage{threeparttable}
\usepackage{color}
\usepackage[ruled,vlined,linesnumbered]{algorithm2e}

\DeclareMathOperator*{\argmin}{arg\,min}
\catcode`\%=12
\newcommand\pcnt{
\catcode`\%=14

\title{LSEC: Large-scale spectral ensemble clustering}

\author{Hongmin Li, Xiucai Ye, Akira Imakura and Tetsuya Sakurai}
\date{%
    \textit{Department of Computer Science} \\
    \textit{University of Tsukuba}\\
    Tsukuba, Japan \\
    li.hongmin.xa@alumni.tsukuba.ac.jp, \{yexiucai, imakura, sakurai\}@cs.tsukuba.ac.jp\\
}
\begin{document}
\maketitle

\begin{abstract}
Ensemble clustering is a fundamental problem in the machine learning field, combining multiple base clusterings into a better clustering result. 
However, most of the existing methods are unsuitable for large-scale ensemble clustering tasks due to the efficiency bottleneck. 
In this paper, we propose a large-scale spectral ensemble clustering (LSEC) method to strike a good balance between efficiency and effectiveness.
In LSEC, a large-scale spectral clustering based efficient ensemble generation framework is designed to generate various base clusterings within a low computational complexity.
Then all based clustering are combined through a bipartite graph partition based consensus function into a better consensus clustering result.
The LSEC method achieves a lower computational complexity than most existing ensemble clustering methods.
Experiments conducted on ten large-scale datasets show the efficiency and effectiveness of the LSEC method.
The MATLAB code of the proposed method and experimental datasets are available at \textit{https://github.com/Li-Hongmin/MyPaperWithCode}.
\end{abstract}

\section{Introduction}

Ensemble clustering, also known as consensus clustering, is a classic problem in machine learning field, aiming to combine multiple base clustering into a better and more consensus clustering \cite{strehl2002cluster,li2004combining,fred2005combining,iam2010link,vega2011survey,naldi2013cluster,wu2014k,huang2017locally,huang2018enhanced,liu2017entropy,liu2017spectral,huang2015combining,huang2016ensemble,zheng2014framework,zhong2015clustering}.
Due to its good performance, ensemble clustering has a pivotal role in many research areas, such as community detection \cite{tandon2019fast} and bioinformatics \cite{kiselev2017sc3,wang2017pooled}. 

There are two critical steps in ensemble clustering: ensemble generation and consensus function.
Ensemble generation aims to generate multiple base clusterings on the same datasets.
In the early stage, $k$-means based ensemble generation methods \cite{iam2010link,topchy2003combining,liu2017spectral} are widely used.
Recently, spectral clustering based ensemble generation methods \cite{huang2019ultra,li2020ensemble} have received attention for its high performance.
On the other hand, the consensus function is used to integrate multiple base clusterings into a consensus one.
We can roughly categorize ensemble clustering according to the consensus function into two categories:
the co-association matrix based methods and the graph partitioning based methods.

The co-association matrix based ensemble clustering method \cite{fred2005combining,iam2010link,wang2009clustering,wang2009generalized} is one of the most widely used ensemble clustering strategies.
A typical example is the evidence accumulation clustering method \cite{fred2005combining}, which counts the frequency of the pair-wise co-occurrence of the same cluster between a pair data points according to base clusterings.
After treating the co-association matrix as a similarity matrix, the hierarchical agglomerative clustering algorithm is applied to obtain the consensus clustering.
Iam-On et al. \cite{iam2010link} extend the EAC method by constructing the co-association matrix based on the common neighborhood information between clusters.
Tao et al. \cite{tao2016robust} propose a robust spectral ensemble clustering method to learns a robust representation for the co-association matrix by capturing the noises and conduct spectral clustering to obtain consensus clustering.
Huang et al. \cite{huang2018enhanced} also enhance the co-association matrix based on similarity mapping from the cluster-level to the object-level and achieve ensemble clustering via fast propagation of cluster-wise similarities.
However, the co-occurrence matrix based methods often lead to high computational cost, which has become a bottleneck for large-scale clustering tasks.
Therefore, most co-association matrix based methods can work well in small-scale datasets but hardly complete large-scale clustering tasks in an acceptable time.

Graph partitioning based ensemble clustering methods \cite{strehl2002cluster,fern2004solving,huang2015robust,li2020ensemble} aim to transform the ensemble clustering problem into a graph partitioning problem to find the consensus clustering. 
Strehl and Ghosh \cite{strehl2002cluster} construct a hypergraph representation by exploring base clusterings and propose three graph partitioning based ensemble clustering methods. 
Huang et al. \cite{huang2015robust} develop a sparse graph with a small number of probably reliable links from base clusterings and find the consensus clustering based on probability trajectory analysis.
Li et al. \cite{li2020ensemble} apply spectral clustering method as base clusterings and take the graph Laplacian matrices of base clusterings as input, then learn a consensus representation by optimizing the graph Laplacians of consensus clustering and base clusterings simultaneously, finally conduct spectral clustering to obtain consensus clustering. 
Although graph partitioning based methods have successfully improved clustering quality, they still have limitations regarding large-scale datasets.

Recently, a few studies have made progress in the application of large-scale data for ensemble clustering.
Wu et al. \cite{wu2014k} propose a $k$-means based consensus clustering (KCC) method, which applies the $k$-means method on a contingency matrix from base clusterings to obtain the consensus clustering result efficiently. 
Liu et al. \cite{liu2017spectral} transform the spectral clustering of the co-association matrix into a weighted $k$-means method and prove two approaches are equivalent, which achieve high efficiency for ensemble spectral clustering.
Huang et al. \cite{huang2019ultra} point out the efficient bottleneck of $k$-means based ensemble generation and apply a large-scale spectral clustering method to fast product the base clusterings, then conduct bipartite graph partitioning to obtain the consensus clustering.
Although these studies have achieved success in their respective fields, 
the large-scale ensemble clustering problem is still a significant challenge due to its high computational complexity, 
Moreover, it is noteworthy that the ensemble generation step considerably takes up the run-time during large-scale ensemble clustering tasks, which has been rarely investigated in the literature.

In light of this, we propose a large-scale ensemble spectral clustering (LSEC) method to alleviate the problem of the application of ensemble clustering for large-scale data.
In LSEC, a spectral clustering based ensemble generation method is designed to handle nonlinear datasets efficiently and provide high-quality base clusterings. 
The ensemble generation process is further accelerated by reusing $K$-nearest neighbors among base clusterings and using light-$k$-means to obtain the clustering results.
After ensemble generation, a bipartite graph between data points and clusters from base clusterings is constructed to produce consensus clustering through the bipartite graph partitioning method efficiently.
Experimental results on ten large-scale data sets demonstrate that LSEC delivers highly efficient and high-quality clustering performance compared to some state-of-the-art consensus clustering methods. 

The main contributions of the proposed method are summaries as follows: 

\begin{itemize}
  \item An efficient spectral clustering based ensemble generation method is designed to handle large-scale datasets and provide high-quality base clusterings via divide-and-conquer based large-scale spectral clustering method. 
  \item Two accelerating tricks are proposed: 1) the computation of similarity among multiple base clusterings is accelerated by reusing the $K$-nearest neighbors; 
  2) the process of obtaining base clustering results is accelerated by the light-$k$-means method. 
  \item The proposed method efficiently generates base clusterings and conducts bipartite graph partitioning to find the consensus clustering. Its computational and space complexity is dominated by $O(\frac{m}{q}N\alpha d)$ and $O(NK)$, which achieves a lower computational complexity than most existing ensemble clustering methods.
  
\end{itemize}

\section{Preliminaries}
\label{sec:related_work}

\subsection{Ensemble Clustering}
Ensemble clustering aims to combine multiple base clustering algorithms to achieve better clustering results. Let ${X}$ be a dataset ${X}=\left\{x_{1}, \ldots, x_{n}\right\}$ with $n$ data points.
The ensemble generation is the first step, which applies a specific clustering algorithm to produce $m$ base clusterings. 
Let \({\Pi}=\left\{\pi_{1}, \ldots, \pi_{m}\right\}\) be a set of base clusterings, where $pi_i$ is $i$-th base clustering and  \(\pi_{i}=\left\{\pi_{i}\left(x_{1}\right), \pi_{i}\left(x_{2}\right), \cdots, \pi_{i}\left(x_{n}\right)\right\}\) indicates the clustering labels for all data points.
Many studies \cite{fred2005combining,iam2010link,wu2014k,liu2015spectral,tao2016robust,huang2015robust} use $k$-means based ensemble generation while some studies \cite{huang2019ultra, li2020ensemble} points out that spectral clustering based ensemble generation can significantly improve clustering quality on the nonlinear datasets.
After ensemble generation, the consensus function is used to integrate all base clusterings into a consensus one, which is the second step. 

\subsection{Divide-and-conquer based large-scale spectral clustering algorithm}
\label{sec:dnc-sc}
Divide-and-conquer based large-scale spectral clustering algorithm (DnC-SC) has been proposed as an effective method for large-scale clustering tasks \cite{Li2021}. It first constructs an approximate similarity matrix via a divided-and-conquer based landmark selection and approximates $K$-nearest landmark searching. Then, it transfers the original spectral clustering problem into a bipartite graph partition problem to find the low-dimensional embedding by solving a smaller eigen-problem. Finally, it applies $k$-means on the low-dimensional embedding to obtain the final clustering result. 

Let ${R}=\{r_1, r_2,\cdots, r_p\}$ denote a set of landmarks, where $r_i \in \mathbb{R}^d$ has the same dimension as $x_i$. 
The divided-and-conquer based landmark selection is designed to generate a set of landmark points which can best represent the original data $X$. The objective function \eqref{eq:objLandmarkSelection} measure how well $R$ represent $X$ by compute the residual sum of squares (RRS) between each $x_j$ and its nearest $r_i$.
\begin{equation}
  \centering
  \label{eq:objLandmarkSelection}
  \begin{split}
    \zeta = \sum_{i = 1}^p\sum_{x_j\in S_i}\left\|x_j-r_i\right\|^{2},\\
  \end{split}
\end{equation}
where $\zeta$ denotes RSS and $S_i,S_2,\dots,S_p$ indicate the subsets that are nearest to $r_{1}, r_{2}, \cdots, r_{p}$, respectively.
For each subset $S_i$, $r_{i}$ is the subset center. The objective function \eqref{eq:objLandmarkSelection} can be rewritten as follows: 
\begin{equation}
    \label{eq:objFuncLS}
    g(X, p)= \argmin_{S_1, \dots, S_p} \sum_{i = 1}^p\sum_{x_j\in S_i}\left\|x_j-r_i\right\|^{2}.
\end{equation}
The recursive functions \eqref{eq: ai} and \eqref{eq: gm} are used to divide the optimization problem into small sub-problems which are easier to be solved.
The parameter $\alpha $ is used to determine the upper bound of $k_i$, which controls the landmark selection rate.
\begin{align}
    &g(Q,h) = \bigcup_{i =1}^m g(A_i, k_i), \label{eq: ai}\\
    &\{A_1, \dots, A_m\} = g(Q,m), \label{eq: gm}
\end{align}
The light-$k$-means algorithm \cite{Li2021} is used to solve the larger dividing process $g(\cdot)$ (with more than $10p$ samples), which randomly selects a part of samples to find subset by $k$-means and then assign remained data points to the nearest subsets.
For the smaller dividing processes (with less than or equal to $10p$ samples), $k$-means is directly used to find the subsets.

The similarities between each $x_i\in X$ and its $K$-nearest landmarks are used to construct a sparse similarity matrix.
The centers' nature of landmarks is used to estimate the $K$-nearest landmarks.
Let $S_{x_i}$ be the subset and $x_i \in S_{x_i}$. Denote $r^1_{x_i}$ is the landmark that is the center of $S_{x_i}$. According to the center's nature of landmark, $r^1_{x_i}$ is treated as the nearest landmark of $x_i$.
In DnC-SC, a set of $K'$-nearest landmarks ($K'>K$) of $r^1_{x_i}$ is first obtained, denoted as ${N}_{K'}(r_{x_i}^{1})$;
then $K$-nearest landmark of $x_i$ are searched from ${N}_{K'}(r_{x_i}^{1})$, denoted as ${N}_{K}(x_i)$.
Finally, the sparse similarity matrix $B$ is constructed as follows \cite{ye2016robust,ye2018spectral}:
\begin{align}
  b_{ij} & = \begin{cases}\exp( \frac{-\left \| x_i -r_j \right \|^2}{2\sigma^2})
    ,  & \text{if $r_j\in N_K(x_i)$},\label{eq:gaussian kernel}
    \\
    0, & \text{otherwise,}
  \end{cases}
\end{align}
where the Gaussian kernel is used to measure the similarity and $\sigma$ is the bandwidth parameter.

The similarity matrix $B$ reflects the relationship between data $X$ and landmarks $R$, which can be treated as the edge of the bipartite graph $G(X, R, B)$.
Therefore, the spectral clustering problem is converted into a bipartite graph partition problem.
According to \cite{Li2021}, the low-dimensional embedding of $R$ side can be computed as follows:
\begin{align}
  \label{eq:general_eigen_problem_reduced}
  L_{{R}}V=\lambda D_{{R}}V,\\
  U = D_{X}^{-1}BV. \label{eq:u_in_practice}
\end{align}
where $L_{{R}}=D_{{R}}-B^TD_{X}^{-1}B$, $D_{X} \in \mathbb{R}^{n \times n}$ and $D_{{R}}\in \mathbb{R}^{p \times p}$ are the diagonal matrices whose entries are $d_{X}(i,i) = \sum_{j=1}^n B_{ij}$ and $d_{{R}}(j,j) = \sum_{i=1}^n B_{ij}$, respectively.
\eqref{eq:general_eigen_problem_reduced} is a small eigen-problem with size $p\times p$. $U$ is the $c$ bottom eigenvectors of $X$ side.
Finally, $k$-means is conducted on $U$ to find $c$ clusters as the final clustering result.

\section{Proposed Framework}

\begin{figure*}[tb]
    \centering
    \includegraphics[width=1.0\textwidth]{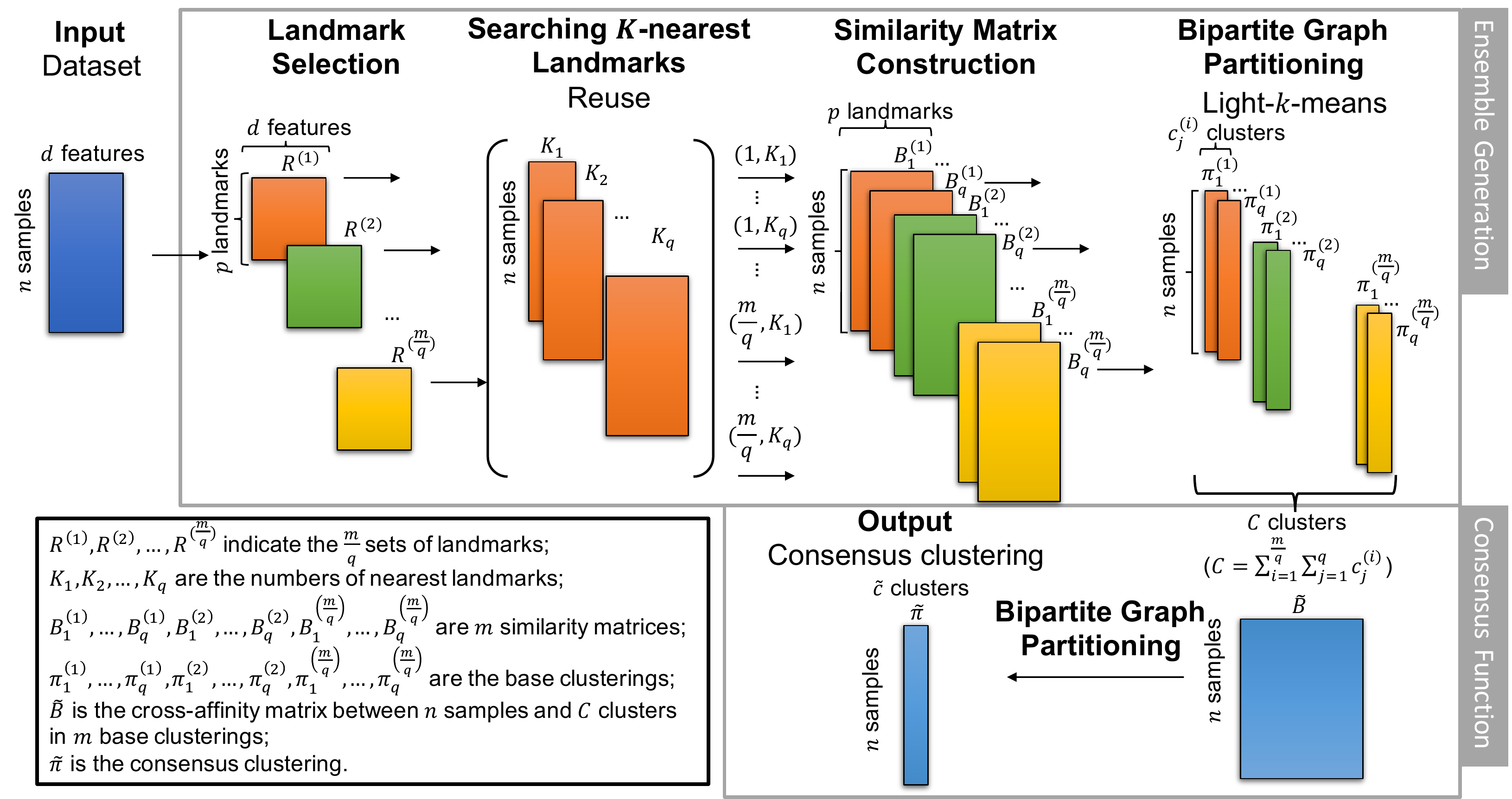}
    \caption{An overview of proposed method. Given a dataset, $\frac{m}{q}$ sets of landmarks are first generated, then a set of $K$-nearest neighbors are found for each $R^{(i)}$ and $m$ sparse similarity matrices are constructed, finally the base clusterings are obtained through a bipartite graph partitioning process.  
    The proposed method accelerates the similarity matrix construction by recycling $K$-nearest neighbors and bipartite graph partitioning by applying light-$k$-means.}
    \label{fig:ensemble_generation}
\end{figure*}

To improve the scalability of ensemble clustering, we propose the LSEC method that complies with the large-scale spectral clustering based formulation and aims to break through the efficiency bottleneck of previous algorithms.
LSEC method consists of two steps: (1) Large-scale spectral clustering based ensemble generation: we design a new framework that applies the state-of-the-art large-scale spectral clustering algorithm to product base clusterings and further accelerate the process by reusing the $K$-nearest landmarks and using light-$k$-means to obtain base clustering results. 
(2) Bipartite graph partitioning based consensus function: we construct a bipartite graph between data points and clusters from base clusterings and obtain the consensus clustering result by bipartite graph partitioning.
Fig.~\ref{fig:ensemble_generation} shows an overview of proposed method.

\subsection{Ensemble Generation based on Large-scale Spectral Clustering}
The ensemble generation step aims to produce diverse $m$ base clusterings with high efficiency.
To improve the scalability of ensemble generation, we consider the divide-and-conquer based large-scale spectral clustering \cite{Li2021} as the base clustering algorithm, which can better handle nonlinear datasets than transitional clustering algorithm like $k$-means and maintain high efficiency.
For better diversity of base clusterings and higher efficiency, we construct similarity matrices with multiple $K$-nearest neighbors graph of sparsification via reusing the $K$-nearest landmarks.
Moreover, the bipartite graph partitioning is accelerated by applying light-$k$-means to obtain the clustering results.

\label{sec:ensemble_generation}

\subsubsection{Landmark Selection}
First, the $\frac{m}{q}$ sets of landmarks are independently generated by solving the optimization problem \eqref{eq:objFuncLS}. 
We recursively apply \eqref{eq: ai} and \eqref{eq: gm} to find an approximate local solution and turn the subset centers as landmarks.
Let ${R}^{(i)}=\{r_1^{(i)}, r_2^{(i)}, \cdots, r_p^{(i)}\}$ is a set of landmarks.
Repeat the divide-and-conquer based landmark selection $\frac{m}{q}$ times, we have $\frac{m}{q}$ sets of landmarks as follows:
\begin{align}
\mathcal{R} = \{{R}^{(1)}, {R}^{(2)}, \cdots, {R}^{(\frac{m}{q})}\},
\end{align}
where ${R}^{(i)}$ indicates the $i$-th set of landmarks and ${R}$ is a set containing all ${R}^{(i)}$, $i=1,2,...,\frac{m}{q}$.
The generation of each ${R}^{(i)}$ costs $O(N\alpha d)$ time complexity and constructing  $\mathcal{R}$ totally costs $O(\frac{m}{q}N\alpha d)$ time complexity.

\subsubsection{Searching $K$-nearest landmarks}

To construct a sparse similarity matrix with $K$-nearest neighbor sparsification, we need to search the $K$-nearest landmarks for each data point $x_i$.
We consider constructing multiple similarity matrices with different sparsification of $K$-nearest landmarks for better diversity of base clusterings. 
Let $K_1 < K_2 < \dots < K_q$ be a set of numbers. 
We search $K_1, K_2, \dots, K_q$-nearest landmarks for each $x_i$, denoted as ${N}_{K_1}(x_i), {N}_{K_2}(x_i), \dots, {N}_{K_q}(x_i)$.
According to the definition of $K$-nearest neighbors, we have
\begin{equation}
\label{eq:subsets}
    {N}_{K_1}(x_i) \subset {N}_{K_2}(x_i) \subset \dots \subset {N}_{K_q}(x_i).
\end{equation}
That is, $N_{K_{j_1}}(x_i)$ is a subset of $N_{K_{j_2}}(x_i)$ if $K_{j_1} < K_{j_2}$.
Therefore, we only need to compute $K_q$-nearest landmarks and then obtain the other $K_1, K_2, \dots, K_q$-nearest landmarks based on it without recomputing. We call this process reusing the nearest landmarks.
Reusing the nearest landmarks accelerates the process of spectral clustering based ensemble generation.
It directly reduces the computational time in two high-cost steps, landmark selection and searching $K$-nearest landmarks, by nearly $q$ times.
Besides efficiency, it also enhances the diversity of base clusterings by exploring multiple nearest neighbor graphs, which is helpful to improve the effectiveness of the proposed method.

\subsubsection{Similarity Matrix Construction}

Then, the sparse similarity matrix between $X$ and each ${R}^{(i)}$ is constructed according to \eqref{eq:gaussian kernel}.
Instead of constructing one similarity matrix for one set of landmarks, we build multiple similarity matrices using different sets of landmarks. 
For each ${R}^{(i)}$, we construct $q$ sparse similarity matrices with $K_1, K_2, \dots, K_q$-nearest landmarks according to (\ref{eq:gaussian kernel}), respectively.
We construct $m$ similarity matrices as follows:
\begin{equation}
\mathcal{B} = \{{B}^{(1)}_1,\dots, {B}^{(1)}_q,{B}^{(2)}_1,\dots,{B}^{(2)}_q,
\dots,{B}^{(\frac{m}{q})}_1,\dots,{B}^{(\frac{m}{q})}_q\},
\label{eq：Bset}    
\end{equation}
where ${B}^{(i)}_j$ indicates a similarity matrix between $X$ and $R^{(i)}$ with sparsification of $K_j$-nearest landmarks, $\mathcal{B}$ is a set containing all ${B}^{(i)}_j$ and the total size of $\mathcal{B}$ is $m$.
The computational cost to obtain a sparse similarity matrix ${B}^{(i)_j}$ is $O(NK_jd)$ \cite{Li2021}.
By reusing the nearest landmarks, we can generate $\mathcal{B}$ with only $O(\frac{m}{q}NK_qd)$ computational cost.
For convenience, we will use $K$ instead of $K_q$ to show the computational complexity in the rest of the paper.

\subsubsection{Bipartite Graph Partitioning}
After obtaining $m$ similarity matrices, we treat each ${B}^{(i)}_j$ as the edge of a bipartite graph $G(X,R^{(\lceil \frac{i}{q} \rceil)},{B}^{(i)_j})$ and solve a bipartite graph partition problem by \eqref{eq:general_eigen_problem_reduced} and \eqref{eq:u_in_practice} to construct a $c^{(i)}_j$-dimensional embedding denoted as ${U}^{(i)_j}$.
Note that $c^{(i)}_j$ is also the number of clusters.
It costs $O(p^3)$ time complexity to solve each bipartite graph partition problem \eqref{eq:general_eigen_problem_reduced} and $O(NK(K+c^{(i)}_j))$ to compute the each $c^{(i)}_j$-dimensional embedding.
The cluster number of $c^{(i)}_j$ is randomly selected as follow:
\begin{align}
c^{(i)}_j=\lfloor\tau(c_{max}-c_{min})\rfloor+c_{min},
\end{align}
where $\tau\in[0,1]$ is a random variable and $c_{max}$ and $c_{min}$ are the upper and lower bounds of the cluster number, respectively. 

The obtained $c^{(i)}_j$ eigenvectors are stacked to form a new matrix, upon which the light-$k$-means \cite{Li2021} is applied to construct the base clustering result. 
In light-$k$-means, a set of $p'$ samples are first randomly selected as representatives, then $c$ clusters centers are generated by applying $k$-means clustering on $p'$ representatives, finally, assign labels to remained samples according to their nearest cluster centers.
The computational complexity of light-$k$-means is $O(pcdt+ Ncd)$, where $O(Ncd)$ is the dominated term and $d$ is the dimensional size.
The light-$k$-means alleviates the computational cost from $t$ iterations and can achieve more efficiency on the platform optimized for matrix operation.
The use of light-$k$-means significantly accelerates the process of obtaining base clusterings for large-scale datasets.
Finally, $m$ base clusterings are generated, which are represented as
\begin{align}
\label{eq: CCC }
\Pi=\{{\pi}^{(1)}_1,\dots, {\pi}^{(1)}_q,{\pi}^{(2)}_1,\dots,{\pi}^{(2)}_q,
\dots,{\pi}^{(\frac{m}{q})}_1,\dots,{\pi}^{(\frac{m}{q})}_q\},
\end{align}
where $\pi^{(i)}_j$ denotes a base clustering with $c^{(i)}_j$ clusters.
For convenience, we use $c$ instead of $c^{(i)}_j$ to show the computational complexity in the rest paper. 
The computational complexity of using light-$k$-means is $O(Nc^2+ p'c^2t)$, where $O(Nc^2)$ is the dominated term. 
Overall, the computational complexity of the bipartite graph partition is $O(m(N(K^2+c^2 +Kc) +p^3))$.
We summarize the ensemble generation process of the proposed method in algorithm~\ref{ag:algorithm1}.

\begin{algorithm}
  
  \caption{Proposed ensemble generation}
    \label{ag:algorithm1}
  \SetAlgoLined
  \KwIn{Dataset $X$, number of base clusterings $m$, a set of number of $K$-nearest landmarks $K_1,K_2,\dots,K_q$}
  \KwOut{base clusterings $\pi_1, \pi_2, \dots, \pi_m$}
  
  Solve \eqref{eq:objLandmarkSelection} by recursively applying \eqref{eq: ai} and \eqref{eq: gm} to obtain $\frac{m}{q}$ sets of landmarks $\mathcal{R}$;\\
  \For{$i\gets 1$ \KwTo $\frac{m}{q}$}{
    Search $K_q$-nearest landmarks of each data points according to \cite{Li2021};\\
    \For{$j\gets 1$ \KwTo $q$}{
    Obtain $K_j$-nearest landmarks of each data points according to \eqref{eq:subsets};\\
    Construct similarity matrix between $X$ and $R^{i}$ with sparsification of $K_j$-nearest landmarks by \eqref{eq:gaussian kernel};
    }
  }
  Collect all similarity matrices $\mathcal{B}$ by \eqref{eq：Bset};\\
  \For{$i\gets 1$ \KwTo $m$}{
    Find a low-dimensional embedding $U$ by \eqref{eq:general_eigen_problem_reduced} and \eqref{eq:u_in_practice};\\
    Apply light-$k$-means on the embedding $U$ to obtain base clustering $\pi_i$.
    }
\end{algorithm}

\subsection{Consensus Function based on Bipartite Graph Partitioning}
\label{sec:consensus_function}

After ensemble generation, the base clusterings will be combined according to a consensus function for obtaining the consensus partition. Again, we treat this problem as a bipartite graph partition problem and give a similar solution like section \ref{sec:dnc-sc}.
\begin{table}
    \caption{The cluster indicator matrix}
    \centering
    \begin{tabular}{|c|cccc|c|}
\hline & $\omega_{1}$ & $\omega_{2}$ & $\cdots$ & $\omega_{C}$ & $\sum$ \\
\hline$\mathcal{X}_1$ & $\tilde{b}_{11}$ & $\tilde{b}_{12}$ & $\cdots$ & $\tilde{b}_{1 C}$ & $m$ \\
$\mathcal{X}_2$ & $\tilde{b}_{21}$ & $\tilde{b}_{22}$ & $\cdots$ & $\tilde{b}_{2 C}$ & $m$ \\
$\cdot$ & $\cdot$ & $\cdot$ & $\cdots$ & $\cdot$ & $\cdot$ \\
$\mathcal{X}_{n}$ & $\tilde{b}_{n 1}$ & $\tilde{b}_{n 2}$ & $\cdots$ & $\tilde{b}_{n C}$ & $m$ \\
\hline$\sum$ & $\|\omega_{1}\|$ & $\|\omega_{2}\|$ & $\cdots$ & $\|\omega_{C}\|$ & $N m$ \\
\hline
\end{tabular}
    \label{tab:indicator}
\end{table}

To define the bipartite graph, we first collect all clusters though the base clusterings as \eqref{eq: CCC } and we denotes the clusters in \eqref{eq: C} for clarity.
\begin{align}
\Psi=\{{\Omega}^{(1)}_1,\dots, {\Omega}^{(1)}_q,{\Omega}^{(2)}_1,\dots,{\Omega}^{(2)}_q, \dots,{\Omega}^{(\frac{m}{q})}_1,\dots,{\Omega}^{(\frac{m}{q})}_q\}, \label{eq: C}
\end{align}
where $\Omega^{(i)}_j$ indicate the set of clusters in $\pi^{(i)}_j$.
There are $c^{(i)}_j$ clusters in each $\Omega^{(i)}_j$, which we denote as:
\begin{equation}
    \Omega^{(i)}_j = \{\omega'_1, \omega'_2, \dots, \omega'_{c^{(i)}_j}\},
\end{equation}
where $\omega'_t$ is the $t$-th cluster in $\Omega^{(i)}_j$. 
Thus, the total number of clusters in $\Psi$ can be counted as $C=\sum_{i=1}^{\frac{m}{q}}\sum_{j=1}^q c_j^{(i)}$. 
For convenience, we simplify the notation of \eqref{eq: omega} as follows:
\begin{equation}
    \Psi=\{\omega_1, \omega_2, \dots, \omega_C\}, \label{eq: omega}
\end{equation}

After the definition of $\Omega$, we design a bipartite graph between data points and clusters as follow: 
\begin{align}
\tilde{G}=\{\mathcal{X},\Omega,\tilde{B}\},
\end{align}
where $\tilde{B}$ is the cross-affinity matrix between $\mathcal{X}$ and $\Omega$. 
$\tilde{B}$ can also be interpreted as the cluster indicator matrix of ${X}$. Table \ref{tab:indicator} shows the cluster indicator matrix, where $b_{ij}=1$ indicates that ${X}_i \in \omega_j$.
$\tilde{G}$ is an unweighted bipartite graph where any edge between node $X_i$ and $\omega_j$ indicates the cluster relationship $X_i \in \omega_j$.
We can give the formula of $\tilde{B}$ as follow:
\begin{align}
\tilde{b}_{ij}  &= \begin{cases} 1
,&\text{if $x_i\in \omega_j$},\label{eq:biparite_indicate_matrix}
\\
0,&\text{otherwise.}
\end{cases}
\end{align}

As Table \ref{tab:indicator} shows, the sum of each rows of $\tilde{B}$ is as the same as number of base clusterings $m$ because there is only one cluster $x_i$ belongs to each base clustering $\pi^{j}$, i.e., $\forall i'\neq j'$, if $\omega_{i'}\in\pi^i$ and $\omega_{j'}\in\pi^i$, then $\omega_{i'}\bigcap \omega_{j'}=\emptyset$.
Though the number of samples in each $\omega_i$ is uncertain, i.e., $\|\omega_i \|$, the total number of non-zeros entries is clearly $Nm$ (see Table \ref{tab:indicator}).

For this modified bipartite graph $\tilde{G}$, we consider a similar partition strategy to what we introduced in Section~\ref{sec:dnc-sc}.
According to \cite{Li2021}, we can write the full similarity of $\mathcal{G}$ as follow
\begin{equation}
\tilde{W}=\left[\begin{array}{ll}
{0} & \tilde{B} \\
\tilde{B}^{T} & {0}
\end{array}\right].
\end{equation}
Then the we have a generalized eigen-problem of $\tilde{G}$ 
\begin{align}
\tilde{L}\tilde{f} = \lambda \tilde{D} \tilde{f},
\end{align}
where $\tilde{L} = \tilde{D}-\tilde{W}$ and $\tilde{D}$ is a diagonal matrix with $\tilde{d_{ii}} = \sum_{j=1}^n\tilde{w_{ij}}$.
According to \eqref{eq:ensemble_eigen_problem_reduced} and \eqref{eq:u_in_practice}, we design a smaller eigen-problem to compute the eigenvetor $\tilde{U}$ in $\mathcal{X}$ side as follows:
\begin{align}
\label{eq:ensemble_eigen_problem_reduced}
{L}_{\Omega}\tilde{V}=\tilde{\lambda} {D}_{\Omega}\tilde{V},
\end{align}
where $L_{\Omega}=\tilde{D}_{\Omega} -\tilde{B}^\top {\tilde{D}_{\mathcal{X}}}^{-1}\tilde{B}$ is the graph Laplacian, $\tilde{D}_{\mathcal{X}} \in R^{n \times n}$ and $\tilde{D}_{\mathcal{R}}\in R^{p \times p}$ are the diagonal matrices whose entries are $\tilde{d}_{\mathcal{X}}(i,i) = \sum_{j=1}^n \tilde{B}_{ij}$ and $\tilde{d}_{\mathcal{R
}}(j,j) = \sum_{i=1}^n \tilde{B}_{ij}$, respectively. 
The size of $L_{\Omega}$ is $C \times C$. Solving the eigen-problem \eqref{eq:ensemble_eigen_problem_reduced} cost $O(C^3)$ computational time.
Substituting $\tilde{V}$ into \eqref{eq:u_in_practice}, we can computer $\tilde{U}$ as follow
\begin{equation}
    \tilde{D} = \tilde{D}_{\mathcal{X}}^{-1}\tilde{B}\tilde{V}. \label{eq:tilde u_in_practice}
\end{equation}
The $\tilde{c}$ bottom eigenvectors $\tilde{U}$ can be computed with with $O(Nm(m+c))$ time.
Finally, the consensus clustering results in LSEC can be obtained by the $k$-means method with $O(Nc^2t)$ time.
We summarize the proposed method LSEC in algorithm~\ref{ag:algorithm2}.


\begin{algorithm}
  \caption{Large-scale ensemble spectral clustering}
  \label{ag:algorithm2}
  \SetAlgoLined
  \KwIn{Dataset $X$, number of base clusterings $m$, a set of number of $K$-nearest landmarks $K_1,K_2,\dots,K_q$, $m$ base clusterings, number of clusters $\tilde{c}$}
  \KwOut{Consensus clustering $\tilde{\pi}$}
  Produce $m$ base clustering by large-scale ensemble generation;\\
  Construct the cluster indicator matrix $\tilde{B}$ according to \eqref{eq:biparite_indicate_matrix};\\
  Solve the eigen-problem \eqref{eq:ensemble_eigen_problem_reduced} to compute $\tilde{V}$;\\
  Find a low-dimensional embedding $\tilde{U}$ of $X$ by \eqref{eq:tilde u_in_practice};\\
  Applying $k$-means to find $\tilde{c}$ clusters on $\tilde{U}$ to obtain consensus clustering.
  Obtain consensus clustering by large-scale consensus function.
\end{algorithm}
\begin{table*}[bp]
\centering
 \caption{Comparison of the computational complexity between LSEC and U-SPEC.}
  \label{table:cmp_complexity}
     \resizebox{\textwidth}{!}{%
\begin{tabular}{@{}lllll@{}}
\toprule
\multirow{2}{*}{Method} & \multicolumn{3}{l}{Ensemble Generation}                                                          & \multirow{2}{*}{Consensus Function} \\ \cmidrule(lr){2-4}
                        & Landmark selection & Similarity construction & Bipartite graph partitioning &                                     \\ \midrule
U-SPEC                  & $O(mp^2dt)$         & $O(mNp^{\frac{1}{2}}d)$  & $O(m(N(K^2+c^2t+Kc)+p^3))$                  & $O(N(m^2+mk+c^2t)+C^3)$           \\
LSEC                    & $O(\frac{m}{q}N\alpha d)$     & $O(\frac{m}{q}NKd) $          & $O(m(N(K^2+c^2+Kc)+p^3))$                   & $O(N(m^2+mk+c^2t)+C^3)$           \\ \bottomrule
\end{tabular}
}
\end{table*}
\section{Discussion}

\subsection{Computational Complexity Analysis}
\label{sec:dnc-sc_complexity}

In this section, we summarize the computational cost of the proposed method.
The ensemble generation of LSEC algorithm takes $O(mN(\alpha d+K^2+Kc+Kd+qc^2)+p^3+p^2(d+K))$ computational cost.
The consensus function of LSEC takes $O(N((qm)^2+qmk+c^2t)+{C}^3)$ time.
With consideration to $m,q,k,K < \alpha \ll p\ll N$, the dominant term of the overall time complexity of LSEC is $O(Nm(\alpha d+qk^2))$.
Meanwhile, the memory costs of the ensemble generation and the consensus function of our LSEC algorithm are respectively $O(N\alpha )$ and $O(Nm)$.
Table~\ref{table:cmp_complexity} provides a comparison of the computational complexity of our DnC-SC algorithm against a state-of-the-art large-scale ensemble clustering method U-SPEC.

\subsection{Relations with Other Methods}
As a large-scale spectral ensemble clustering method, the proposed method is closely related to the U-SENC method in \cite{huang2019ultra}. 
We compare the proposed method with the U-SENC to discuss the improvements of the proposed method. 

Firstly, we compare them to the ensemble generation methods in the term of diversity and efficiency. 
In U-SENC, base clusterings are directly generated using a large-scale spectral clustering U-SPEC with different numbers of clusters.
As a large-scale spectral clustering method, the U-SPEC method also uses the landmark selection technique.
Thus, the diversity of base clusterings of U-SENC is from two facts: the different landmarks and the number of clusters of ensemble generation.
However, the $K$-nearest neighbor graph is not used to improve the diversity in U-SENC further.
In our proposed method, we consider the various landmarks and number of clusters and use different $K$-nearest neighbors to construct a sparse similarity matrix to improve the overall diversity of base clustering. 

Secondly, we compare them to the ensemble generation methods in the term of efficiency.
Since the different $K$-nearest neighbor graphs can share the same $K$-nearest neighbors between data points and landmarks, the computational complexity of similarity matrix construction is much less than the U-SENC method.
For large-scale datasets, another computational bottleneck is the final $k$-means step of large-scale spectral clustering.
In our proposed method, we use the light-$k$-means to accelerate the base clustering results, significantly improving large-scale datasets' efficiency. 

Overall, LSEC redesigns the ensemble generation framework based on a more efficient clustering method (i.e., DnC-SC) and accelerates the process by reusing the $K$-nearest neighbors among multiple base clusterings.
Furthermore, a light-$k$-means method is used to fast obtain the base clustering results.
The computational complexity of the proposed method is faster than most existing large-scale ensemble clustering methods.

\section{Experiments}
\label{sec:experiment}

In this section, we conduct experiments on five real and five synthetic datasets to evaluate the performance of the proposed LSEC method.
The comparison experiments against several state-of-the-art spectral clustering methods show better clustering quality and efficiency for LSEC methods.
Besides that, the analysis of parameters is performed.
For each experiment, the test method is repeated 20 times, and the average performance is reported.
All experiments are conducted in Matlab R2020a on a Mac Pro with 3 GHz 8-Core Intel Xeon E5 and 16 GB of RAM.

\subsection{Datasets and Evaluation Measures}

\begin{table}[t]
  \centering
  \caption{Properties of the real and synthetic datasets.}
  \label{table:datasets}
  \begin{center}
    \begin{tabular}{p{1.2cm}<{\centering}|p{1.3cm}<{\centering}|p{1.5cm}<{\centering}p{1.2cm}<{\centering}p{1.2cm}<{\centering}}
      \toprule
      \multicolumn{2}{c|}{Dataset}         &\#Object     &\#Dimension      &\#Class\\
      \midrule
                                        & \emph{USPS}      & 9298    & 256 & 10 \\
      \multirow{5}{*}{\emph{Real}}      & \emph{PenDigits} & 10,992    & 16  & 10 \\
                                        & \emph{Letters}   & 20,000    & 16  & 26 \\
                                        & \emph{MNIST}     & 70,000    & 784 & 10 \\
                                        & \emph{Covertype} & 581,012   & 54  & 7  \\
      \midrule
      \multirow{5}{*}{\emph{Synthetic}} & \emph{TB-1M}    & 1,000,000   & 2   & 3  \\
                                        & \emph{SF-2M}     & 2,000,000 & 2   & 4  \\
                                        & \emph{CC-5M}     & 5,000,000 & 2   & 3  \\
                                        & \emph{CG-10M}    & 10,000,000 & 2   & 11  \\
                                        & \emph{FL-20M}    & 20,000,000 & 2   & 13  \\
      \bottomrule
    \end{tabular}
  \end{center}
\end{table}

\begin{figure}[tpb]
  \begin{center}
    {\subfigure[\emph{TB-1M} ($0.1\%$)]
      {\includegraphics[width=0.31\columnwidth]{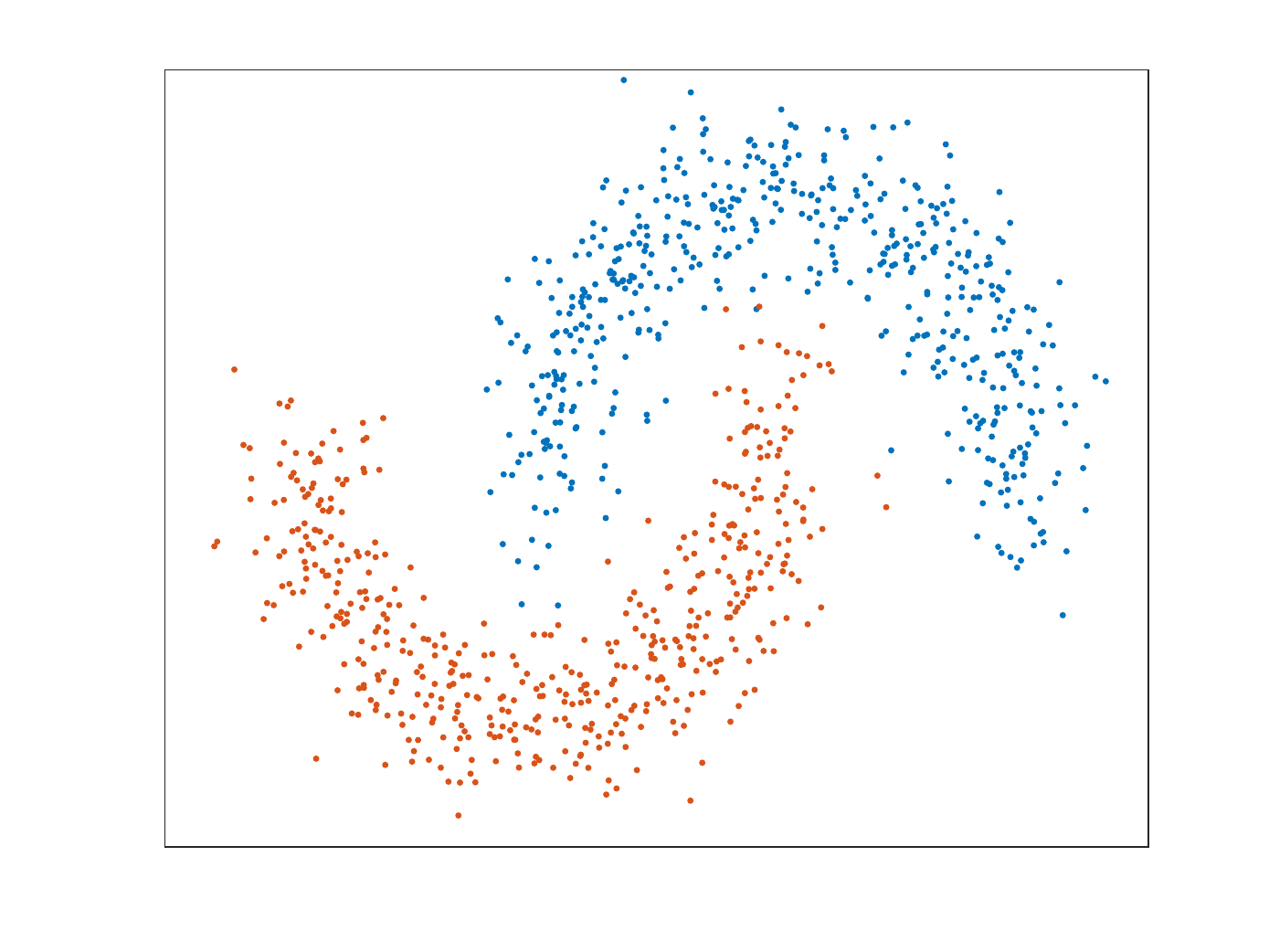}}}
    {\subfigure[\emph{SF-2M} ($0.1\%$)]
      {\includegraphics[width=0.31\columnwidth]{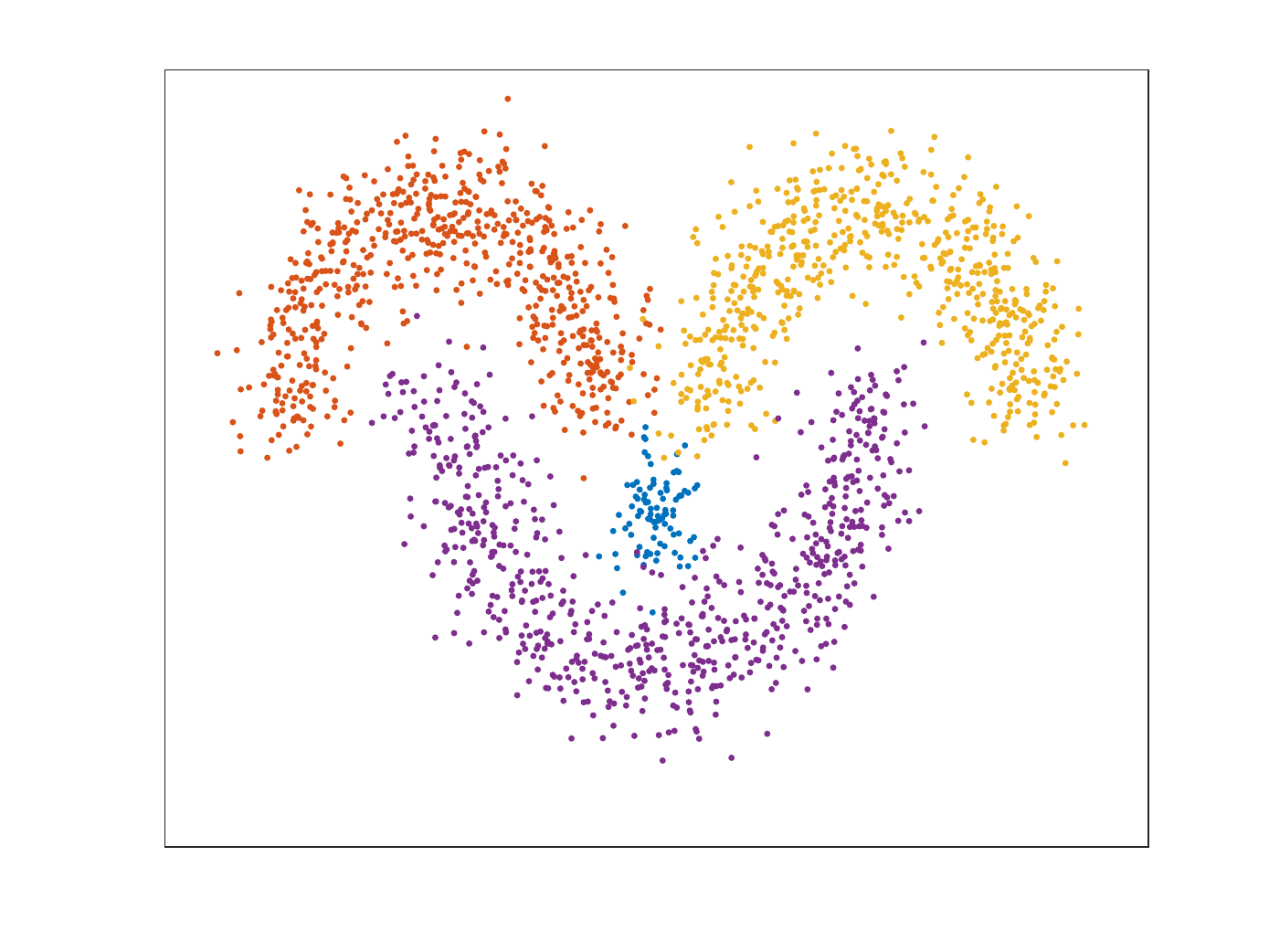}}}
    {\subfigure[\emph{CC-5M} ($0.1\%$)]
      {\includegraphics[width=0.31\columnwidth]{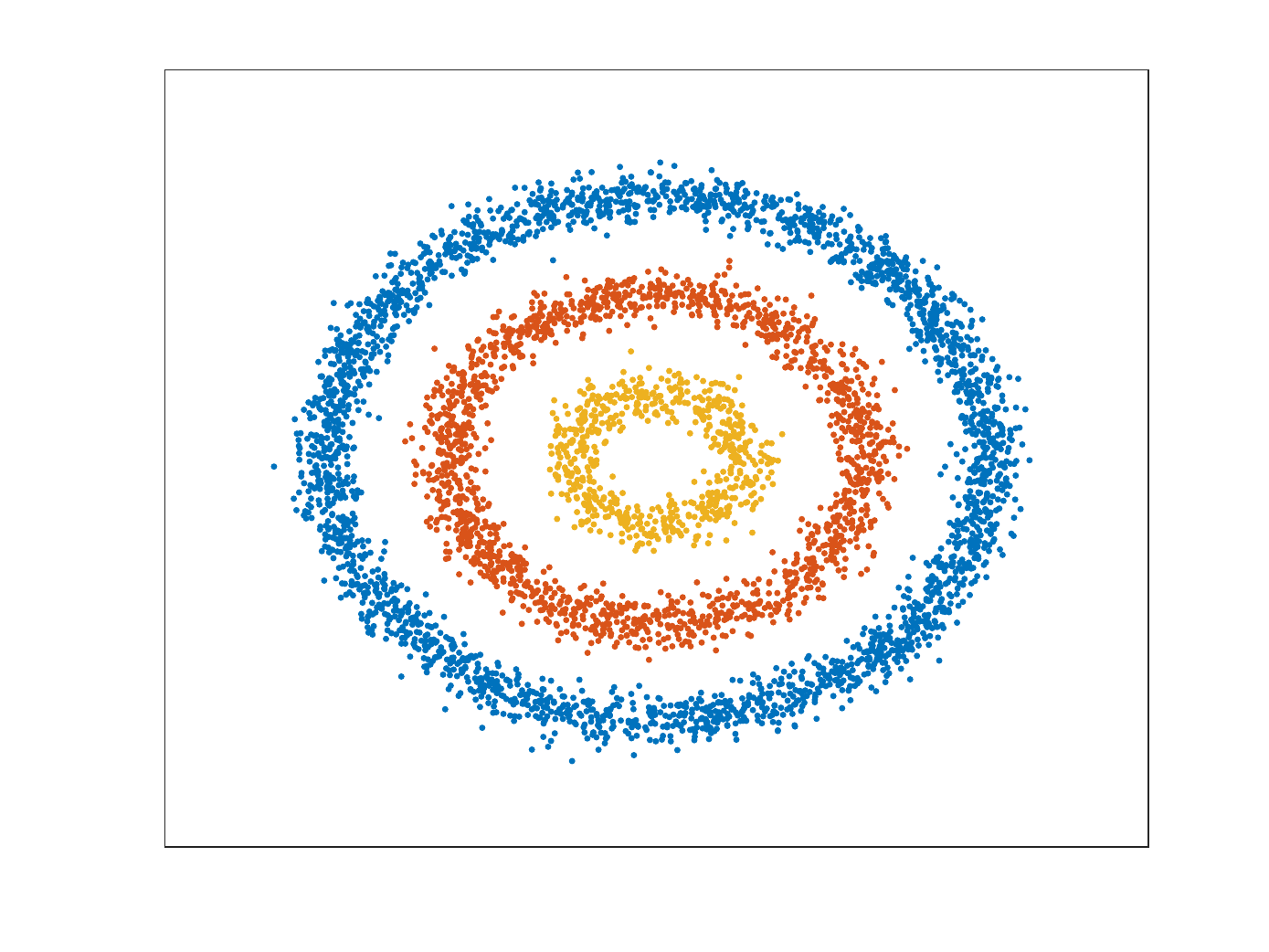}}}
    {\subfigure[\emph{CG-10M} ($0.1\%$)]
        {\includegraphics[width=0.31\columnwidth]{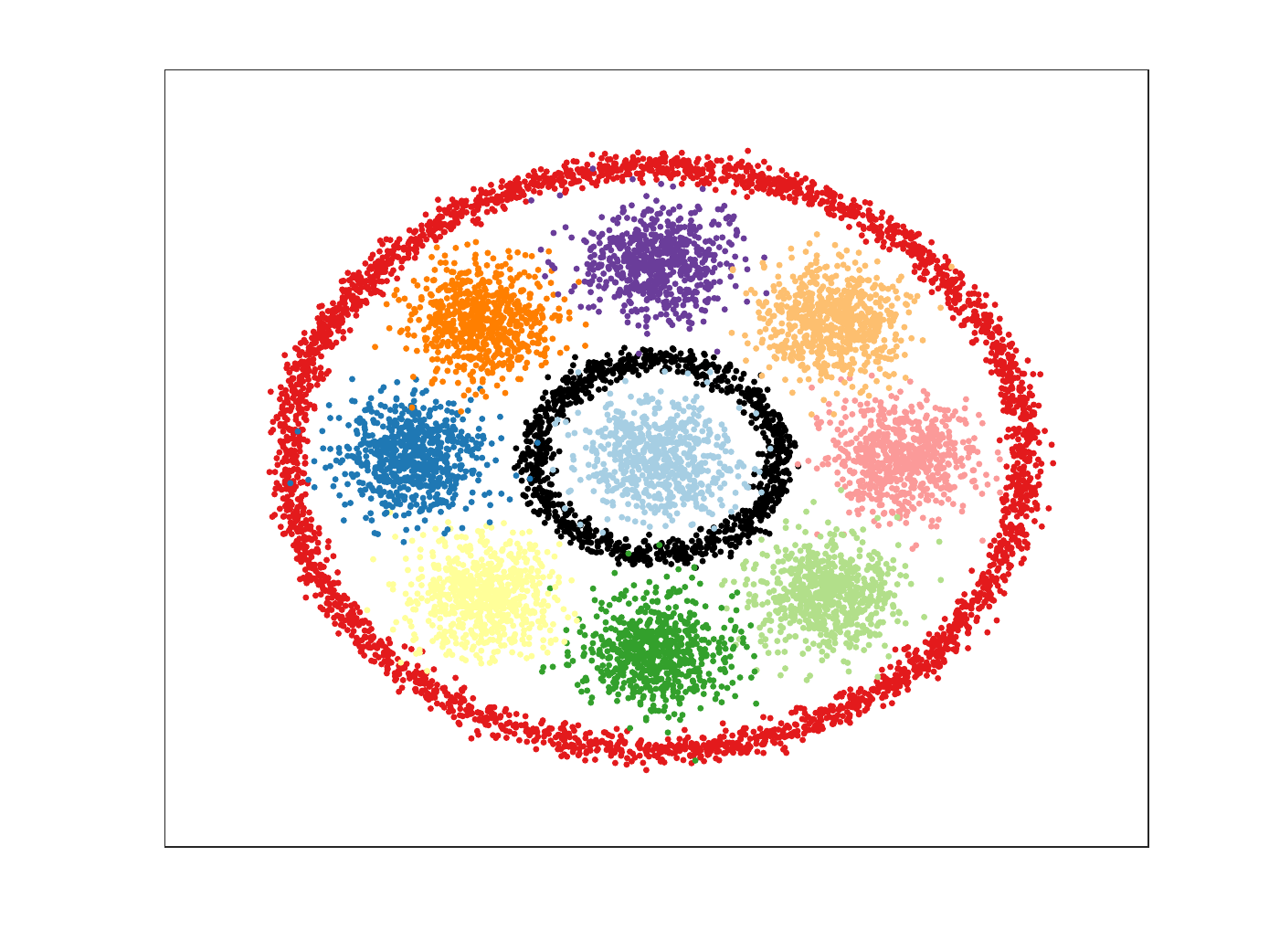}}}
    {\subfigure[\emph{FL-20M} ($0.1\%$)]
        {\includegraphics[width=0.31\columnwidth]{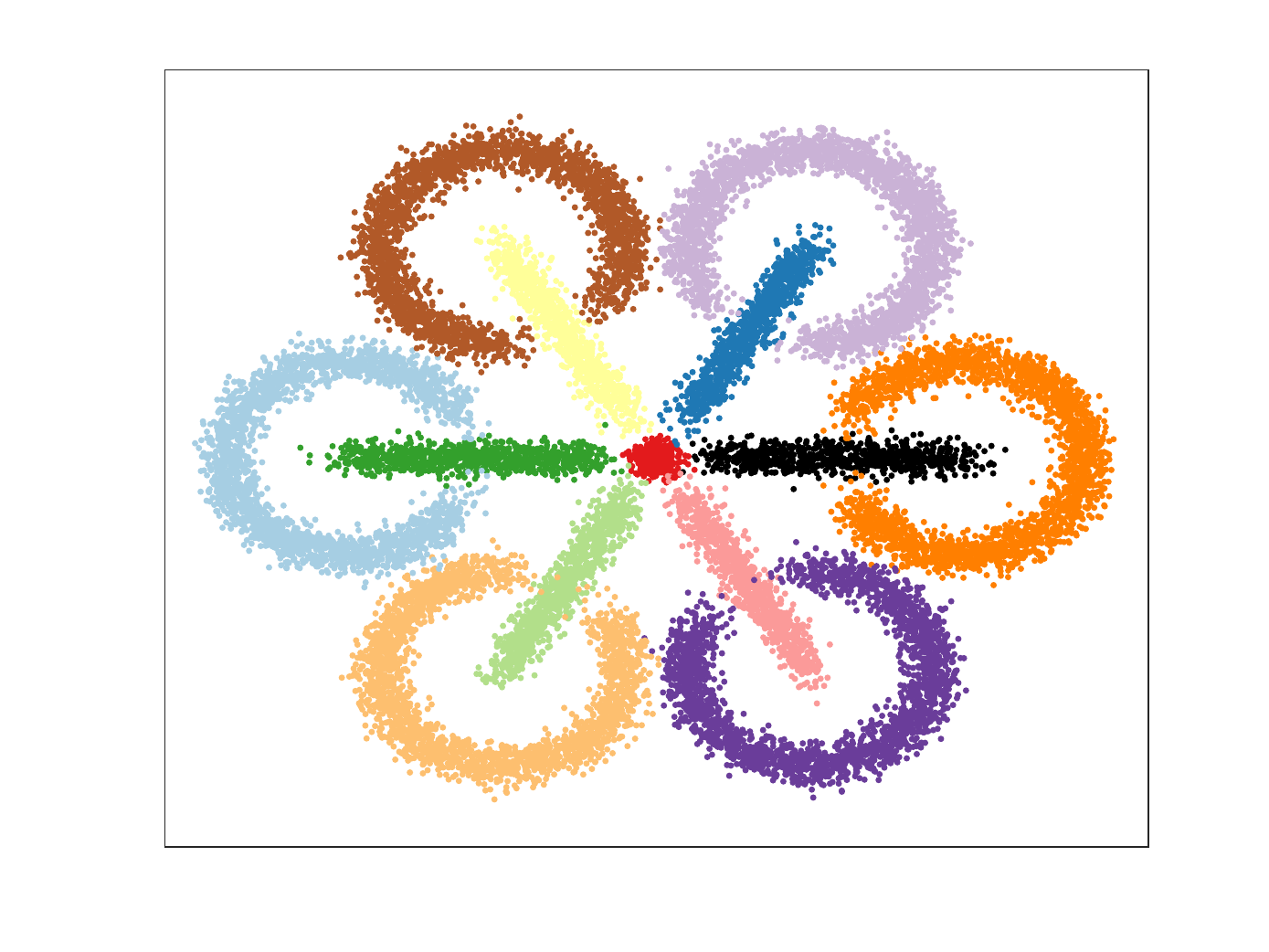}}}
    \caption{Illustration of the five synthetic datasets. Note that only $0.1\%$ samples of each dataset are plotted.}
    \label{fig:fiveSynDS}
  \end{center}
\end{figure}

Our experiments are conducted on ten large-scale datasets, varying from nine thousands to as large as twenty million data points. Specifically, the five real datasets are \emph{PenDigits} \cite{asuncion2007uci} \footnote{https://archive.ics.uci.edu/ml/datasets/Pen-Based+Recognition+of+Handwritten+Digits}, 
\emph{USPS} \cite{cai2010graph} \footnote{\label{cai_deng_data} http://www.cad.zju.edu.cn/home/dengcai/Data/MLData.html}, 
\emph{Letters} \cite{frey1991letter} \footnote{https://archive.ics.uci.edu/ml/datasets/Letter+Recognition}, 
\emph{MNIST} \cite{cai2011speed}, 
and \emph{Covertype} \cite{blackard1999comparative} \footnote{https://archive.ics.uci.edu/ml/datasets/covertype}. 
The five synthetic datasets are \emph{Two Bananas} (\emph{TB-1M}), \emph{Smiling Face-2M} (\emph{SF-2M}), \emph{Concentric Circles-5M} (\emph{CC-5M}), \emph{Circles and Gaussians-10M} (\emph{CG-10M}), \emph{Flower-20M} (\emph{FL-20M}) \cite{huang2019ultra} \footnote{\label{huang}https://www.researchgate.net/publication/330760669}. Fig. ~\ref{fig:fiveSynDS} shows the synthetic datasets.
The properties of the datasets are summarized in Table~\ref{table:datasets}.

We adopt two widely used evaluation metrics, i.e., Normalized Mutual Information (NMI) \cite{slonim2000agglomerative} and Accuracy (ACC) \cite{yan2009fast}, to evaluate the clustering results.
Let $X=[x_1,x_2,...,x_n]$ be the data matrix. For each data point $x_i$, denote $\pi_t(x_i)$ and $\pi_c(x_i)$ as the cluster label of ground truth and obtained cluster label from clustering methods, respectively. The ACC is defined as:
\begin{equation}
  \text{ACC}=\frac{\sum_{i=1}^n \delta(\pi_t(x_i),\text{map}(\pi_c(x_i)))}{n},
\end{equation}
where $n$ is the number of data  and $\delta(\pi_t(x_i),\pi_c(x_i))$ is a function to check $\pi_t(x_i)$ and $\pi_c(x_i)$ are equal or not,  returning 1 if equals otherwise returning 0. The map$(\pi_c(x_i))$ is a best mapping function that maps each predicted label to the most possibly true cluster label by permuting operations \cite{xu2003document}.

The NMI is the normalization of Mutual information by the joint entropy as follow:
\begin{equation}
  \label{NMI}
  NMI= \frac{\sum_{\pi_t(x_i)\in T,\pi_c(x_i)\in C}p(\pi_t(x_i),\pi_c(x_i))\text{ln}\frac{p(\pi_t(x_i),\pi_c(x_i))}{p(\pi_t(x_i))p(\pi_c(x_i))}}
  {-\sum_{\pi_t(x_i)\in T,\pi_c(x_i)\in C}p(\pi_t(x_i),\pi_c(x_i))\text{ln}(p(\pi_t(x_i),\pi_c(x_i)))}
  ,\end{equation}

A better clustering result will provide a larger value of NMI/ACC. Both NMI and ACC are in the range of $[0,1]$.

\subsection{Compared Methods and Experimental Settings}

In this experiments, we compare the proposed method with one baseline clustering method, i.e., the divide-and-conquer based large-scale spectral clustering (DnC-SC) \cite{Li2021}, as well as seven state-of-the-art large-scale spectral clustering methods. The compared spectral clustering methods are listed as follows:

\begin{enumerate}
  \item \textbf{EAC} \cite{fred2005combining}: evidence accumulation clustering.
  \item \textbf{KCC} \cite{wu2014k}: $k$-means based consensus clustering.
  \item \textbf{PTGP} \cite{huang2015robust}: probability trajectory based graph partitioning.
  \item \textbf{SEC} \cite{liu2015spectral}: spectral ensemble clustering.
  \item \textbf{LWEA} \cite{huang2017locally}: locally weighted evidence accumulation.
  \item \textbf{LWGP} \cite{huang2017locally}: locally weighted graph partitioning.
  \item \textbf{U-SENC} \cite{huang2019ultra}: ultra-scalable ensemble clustering.
\end{enumerate}

\begin{table*}[tbp]
  \centering
  \caption{ACC(\%) scores (over 20 runs) by our methods and the baseline ensemble clustering methods (The best score in each row is in bold).}
  \label{table:compare_ens_acc}
     \resizebox{\textwidth}{!}{%
  \begin{tabular}{@{}c||c||cccccc|c@{}}
    \toprule
    Dataset            & DnC-SC                & EAC                & KCC                & PTGP               & SEC                         & LWGP               & U-SENC                      & LSEC                        \\
    \midrule
    \emph{PenDigits }  & 82.27$_{\pm 1.33}$    & 77.67$_{\pm 2.30}$ & 44.68$_{\pm 5.10}$ & 80.89$_{\pm 1.26}$ & 32.37$_{\pm 3.88}$          & 73.66$_{\pm 2.14}$ & 87.02$_{\pm 1.65}$          & \textbf{88.26$_{\pm 2.10}$} \\
    \emph{USPS      }  & 82.55$_{\pm 1.96}$    & 66.76$_{\pm 1.69}$ & 56.37$_{\pm 3.41}$ & 67.14$_{\pm 0.26}$ & 40.77$_{\pm 5.82}$          & 65.82$_{\pm 3.41}$ & 78.25$_{\pm 2.39}$          & \textbf{80.97$_{\pm 5.31}$} \\
    \emph{Letters   }  & 33.54$_{\pm 1.21}$    & 29.79$_{\pm 0.60}$ & 24.46$_{\pm 1.24}$ & 26.66$_{\pm 1.42}$ & 23.60$_{\pm 1.28}$          & 27.88$_{\pm 0.78}$ & \textbf{37.03$_{\pm 1.28}$} & 36.33$_{\pm 0.89}$          \\
    \emph{MINST     }  & 74.24$_{\pm 2.14}$    & N/A                & 45.61$_{\pm 4.96}$ & 66.96$_{\pm 0.68}$ & 33.15$_{\pm 2.07}$          & 56.27$_{\pm 1.47}$ & 75.48$_{\pm 3.01}$          & \textbf{80.19$_{\pm 3.74}$} \\
    \emph{Covertype }  & 23.48$_{\pm 1.86}$    & N/A                & 32.52$_{\pm 0.41}$ & 23.45$_{\pm 0.96}$ & \textbf{39.63$_{\pm 6.15}$} & 30.64$_{\pm 0.42}$ & 21.34$_{\pm 1.06}$          & \textbf{23.42$_{\pm 1.86}$} \\
    \emph{TB-1M      } & 99.62$_{\pm 0.02}$    & N/A                & 67.76$_{\pm 1.41}$ & 81.95$_{\pm 0.00}$ & 67.94$_{\pm 3.66}$          & 99.71$_{\pm 0.45}$ & \textbf{99.75$_{\pm 0.01}$} & 99.72$_{\pm 2.31}$          \\
    \emph{SF-2M      } & 9.43$_{\pm 0.31}$     & N/A                & 50.94$_{\pm 4.15}$ & 60.25$_{\pm 0.94}$ & 49.88$_{\pm 5.68}$          & 80.04$_{\pm 3.45}$ & 76.54$_{\pm 2.88}$          & \textbf{85.17$_{\pm 9.18}$} \\
    \emph{CC-5M      } & 99.98$_{\pm 0.00}$    & N/A                & 72.25$_{\pm 6.41}$ & 34.95$_{\pm 0.00}$ & 41.57$_{\pm 0.81}$          & 97.84$_{\pm 3.71}$ & \textbf{99.99$_{\pm 0.00}$} & 95.61$_{\pm 11.67}$          \\
    \emph{CG-10M   }   & 66.83$_{\pm 4.46}$    & N/A                & 58.12$_{\pm 5.41}$ & 60.89$_{\pm 1.73}$ & 46.26$_{\pm 5.72}$          & 71.95$_{\pm 3.19}$ & 82.34$_{\pm 5.59}$          & \textbf{97.57$_{\pm 3.49}$} \\
    \emph{FL-20M }     & 81.90$_{\pm 5.61}$    & N/A                & 48.21$_{\pm 4.14}$ & 51.21$_{\pm 1.41}$ & 41.70$_{\pm 0.42}$          & 72.15$_{\pm 2.45}$ & 78.16$_{\pm 3.21}$          & \textbf{82.81$_{\pm 3.21}$} \\
    \midrule
    \midrule
    Avg. score         & \multicolumn{1}{c}{-} & N/A                & 50.09              & 55.44              & 41.69                       & 67.60              & 72.59                       & \textbf{77.01}              \\
    \midrule
    \midrule
    Avg. rank          & \multicolumn{1}{c}{-} & 6.00               & 5.00               & 4.00               & 5.60                        & 3.30               & 2.30                        & \textbf{1.90}               \\
    \bottomrule
  \end{tabular}%
     }
\end{table*}
\begin{table*}[tbp]
  \centering
  \caption{NMI(\%) scores (over 20 runs) by our methods and the baseline ensemble clustering methods (The best score in each row is in bold).}
  \label{table:compare_ens_nmi}
     \resizebox{\textwidth}{!}{%
  \begin{tabular}{@{}c||c||cccccc|c@{}}
    \toprule
    Dataset            & DnC-SC                & EAC                & KCC                & PTGP               & SEC                & LWGP               & U-SENC                      & LSEC                        \\
    \midrule
    \emph{PenDigits }  & 82.01$_{\pm 0.21}$    & 76.12$_{\pm 0.00}$ & 53.52$_{\pm 3.46}$ & 78.31$_{\pm 0.34}$ & 46.44$_{\pm 2.10}$ & 76.46$_{\pm 1.43}$ & 83.24$_{\pm 1.11}$          & \textbf{84.65$_{\pm 1.81}$} \\
    \emph{USPS      }  & 82.86$_{\pm 1.08}$    & 69.05$_{\pm 0.00}$ & 58.27$_{\pm 0.25}$ & 70.32$_{\pm 1.02}$ & 49.68$_{\pm 1.89}$ & 70.71$_{\pm 1.46}$ & 82.09$_{\pm 1.65}$          & \textbf{83.51$_{\pm 1.34}$} \\
    \emph{Letters   }  & 45.37$_{\pm 0.85}$    & 39.28$_{\pm 0.00}$ & 34.61$_{\pm 1.40}$ & 36.98$_{\pm 0.99}$ & 32.30$_{\pm 0.89}$ & 39.29$_{\pm 0.46}$ & 46.40$_{\pm 0.20}$          & \textbf{48.71$_{\pm 0.63}$} \\
    \emph{MINST     }  & 72.00$_{\pm 0.51}$    & N/A                & 46.43$_{\pm 4.85}$ & 62.22$_{\pm 1.12}$ & 38.84$_{\pm 1.44}$ & 62.34$_{\pm 0.62}$ & 75.11$_{\pm 0.58}$          & \textbf{79.42$_{\pm 1.45}$} \\
    \emph{Covertype }  & 8.30$_{\pm 0.30}$     & N/A                & 6.38$_{\pm 3.41}$  & 8.25$_{\pm 0.43}$  & 9.23$_{\pm 6.48}$  & 9.06$_{\pm 0.41}$  & 9.34$_{\pm 1.21}$           & \textbf{11.64$_{\pm 1.76}$} \\
    \emph{TB-1M      } & 96.42$_{\pm 0.18}$    & N/A                & 24.54$_{\pm 2.45}$ & 31.89$_{\pm 0.00}$ & 24.74$_{\pm 4.45}$ & 97.16$_{\pm 2.41}$ & \textbf{97.45$_{\pm 0.04}$} & 97.19$_{\pm 9.43}$          \\
    \emph{SF-2M      } & 81.24$_{\pm 0.32}$    & N/A                & 38.06$_{\pm 2.45}$ & 49.74$_{\pm 0.18}$ & 33.65$_{\pm 3.22}$ & 81.95$_{\pm 4.15}$ & 77.57$_{\pm 2.12}$          & \textbf{84.88$_{\pm 6.55}$} \\
    \emph{CC-5M      } & 99.78$_{\pm 0.01}$    & N/A                & 59.24$_{\pm 0.41}$ & 0.13$_{\pm 0.00}$  & 12.93$_{\pm 1.80}$ & 98.15$_{\pm 7.41}$ & \textbf{99.91$_{\pm 0.00}$} & 95.57$_{\pm 10.70}$         \\
    \emph{CG-10M   }   & 80.91$_{\pm 3.59}$    & N/A                & 63.56$_{\pm 0.41}$ & 65.09$_{\pm 0.92}$ & 55.77$_{\pm 6.84}$ & 78.41$_{\pm 2.93}$ & 86.28$_{\pm 2.30}$          & \textbf{95.25$_{\pm 1.32}$} \\
    \emph{FL-20M }     & 87.67$_{\pm 3.18}$    & N/A                & 68.10$_{\pm 2.41}$ & 71.32$_{\pm 1.29}$ & 53.77$_{\pm 2.52}$ & 78.51$_{\pm 1.97}$ & 90.38$_{\pm 2.45}$          & \textbf{91.32$_{\pm 2.44}$} \\
    \midrule
    \midrule
    Avg. score         & \multicolumn{1}{c}{-} & N/A                & 45.27              & 47.43              & 35.74              & 69.20              & 74.78                       & \textbf{77.21}              \\
    \midrule
    \midrule
    Avg. rank          & \multicolumn{1}{c}{-} & 6.30               & 5.40               & 4.30               & 5.80               & 3.00               & 1.90                        & \textbf{1.40}               \\
    \bottomrule
  \end{tabular}%
     }
\end{table*}
\begin{table*}[tbp]
  \centering
  \caption{Time costs(s) of our methods and the baseline ensemble clustering methods.}
  \label{table:compare_ens_time}
     \resizebox{0.9\textwidth}{!}{%
  \begin{tabular}{@{}c||c||cccccc|c@{}}
    \toprule
    Dataset            & DnC-SC                & EAC   & KCC     & PTGP    & SEC           & LWGP    & U-SENC   & LSEC             \\
    \midrule
    \emph{PenDigits }  & 0.64                  & 18.78 & 9.19    & 6.06    & \textbf{3.00} & 4.00    & 18.31    & 3.40             \\
    \emph{USPS      }  & 1.25                  & 25.79 & 23.56   & 41.32   & 15.08         & 15.93   & 28.82    & \textbf{5.81}    \\
    \emph{Letters   }  & 0.90                   & 115   & 48.48   & 89.88   & 10.76         & 11.15   & 20.86    & \textbf{3.71}    \\
    \emph{MINST     }  & 5.11                  & N/A   & 831.12  & 2297.05 & 730.33        & 731.64  & 103.35   & \textbf{21.36}   \\
    \emph{Covertype }  & 13.15                 & N/A   & 634.18  & 16271.2 & 714.86        & 730.33  & 143.41   & \textbf{40.16}   \\
    \emph{TB-1M      } & 5.06                  & N/A   & 984.15  & 849.62  & 693.67        & 709.60  & 265.80   & \textbf{62.50}   \\
    \emph{SF-2M      } & 13.77                 & N/A   & 2225.64 & 1475.08 & 1344.66       & 1566.8  & 623.26   & \textbf{131.67}  \\
    \emph{CC-5M      } & 25.37                 & N/A   & 8541.13 & 3040.33 & 3232.06       & 3006    & 1851.2   & \textbf{321.71}  \\
    \emph{CG-10M   }   & 281.05                & N/A   & 12351.2 & 7244.01 & 7607.84       & 6685.8  & 3561.4   & \textbf{769.51}  \\
    \emph{FL-20M }     & 837.38                & N/A   & 17112.1 & 13343.3 & 14938.73      & 13091   & 11763.07 & \textbf{2396.85} \\
    \midrule
    \midrule
    Avg. score         & \multicolumn{1}{c}{-} & N/A   & 4276.07 & 4465.78 & 2929.10       & 2655.23 & 1837.95  & \textbf{375.65}  \\
    \midrule
    \midrule
    Avg. rank          & \multicolumn{1}{c}{-} & 6.80  & 5.20    & 5.00    & 3.30          & 3.60    & 3.00     & \textbf{1.10}    \\
    \bottomrule
  \end{tabular}%
     }
\end{table*}

There are several common parameters among the methods mentioned above. We set these parameters as follow:
\begin{itemize}
  \item We set the number of landmarks as $p=1000$ for LSEC and U-SPEC. The parameter analysis on $p$ has been conducted in \cite{Li2021}.
  \item We set the $K=5$ for the number of nearest neighbors for LSEC and U-SPEC.
  \item The DnC-SC method has a unique parameter $\alpha $. In the experiments, $\alpha = 50$ is used for all datasets.
  \item The base clusterings are generated by $k$-means or large-scale spectral clustering as suggested by their papers \cite{fred2005combining,wu2014k,huang2015robust,huang2017locally,huang2019ultra}.
  The number of cluster $c$ of base clusterings is randomly selected from $[20,60]$.
  The number of base clusterings is set as $m=20$ for comparison. 
  The parameter analysis on $m$ will be shown in Section~\ref{sec:para_M}.
  \item The true number of classes on each dataset is used to conduct all experiments.
  \item Other parameters in the baseline methods are set as suggested by the original papers.
\end{itemize}

\subsection{Comparison Results}
\label{sec:cmp_ensemble}

\begin{table*}[tpb]
  \centering
  \caption{Clustering performance (ACC(\%), NMI(\%), and time costs(s)) for different methods by varying number of base clusterings $m$.}
  \label{table:compare_para_Msize}
  \begin{threeparttable}
    \begin{tabular}{m{0.08\textwidth}<{\centering}|m{0.2\textwidth}<{\centering}m{0.2\textwidth}<{\centering}m{0.2\textwidth}<{\centering}m{0.2\textwidth}<{\centering}}
      \toprule
      \emph{Dataset} & \emph{MNIST} & \emph{Covertype} & \emph{TB-1M} & \emph{SF-2M} \\
      \midrule
      \multirow{1}{*}{ACC}
      &\includegraphics[width=0.2\textwidth]{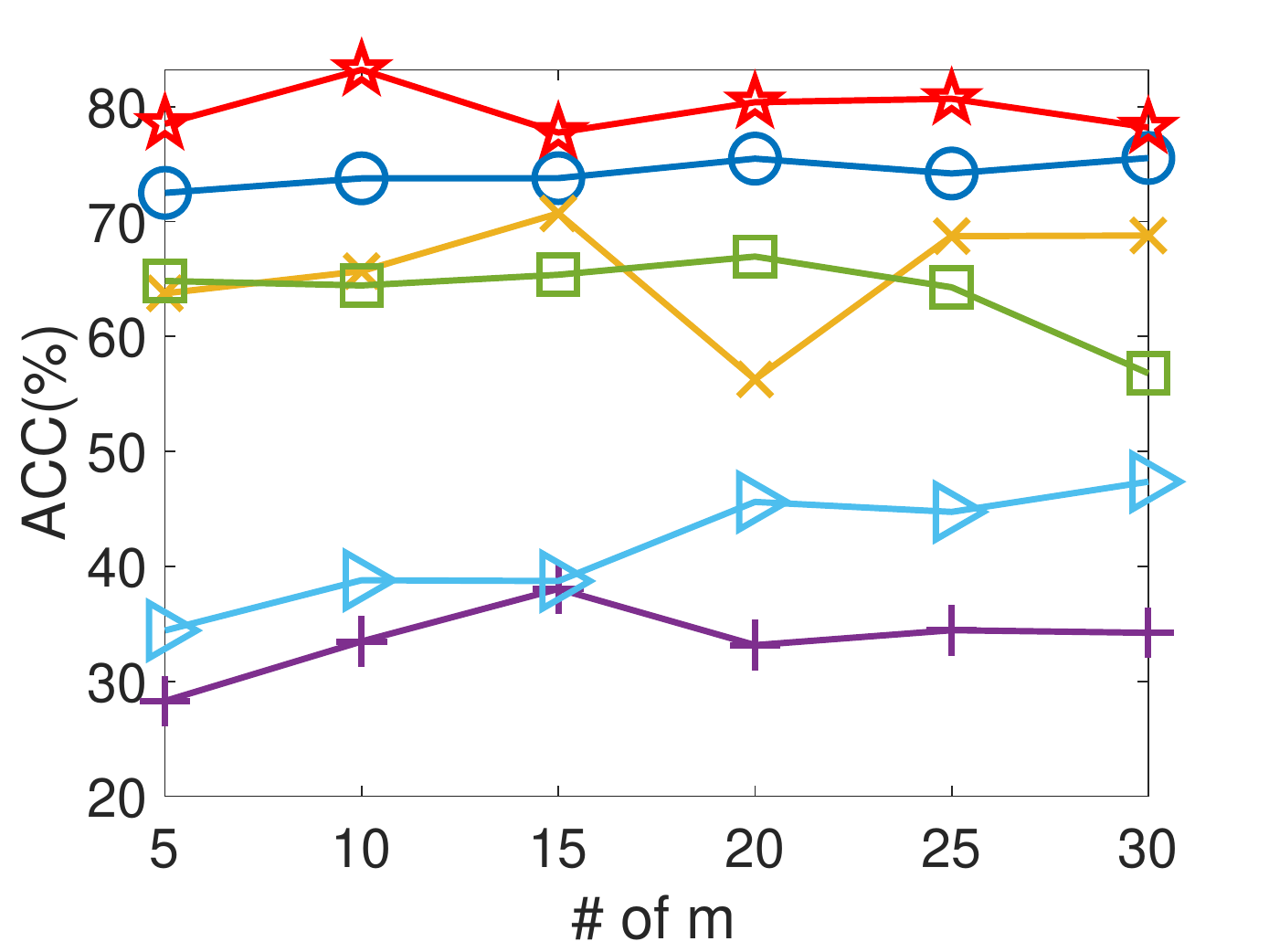}
      &\includegraphics[width=0.2\textwidth]{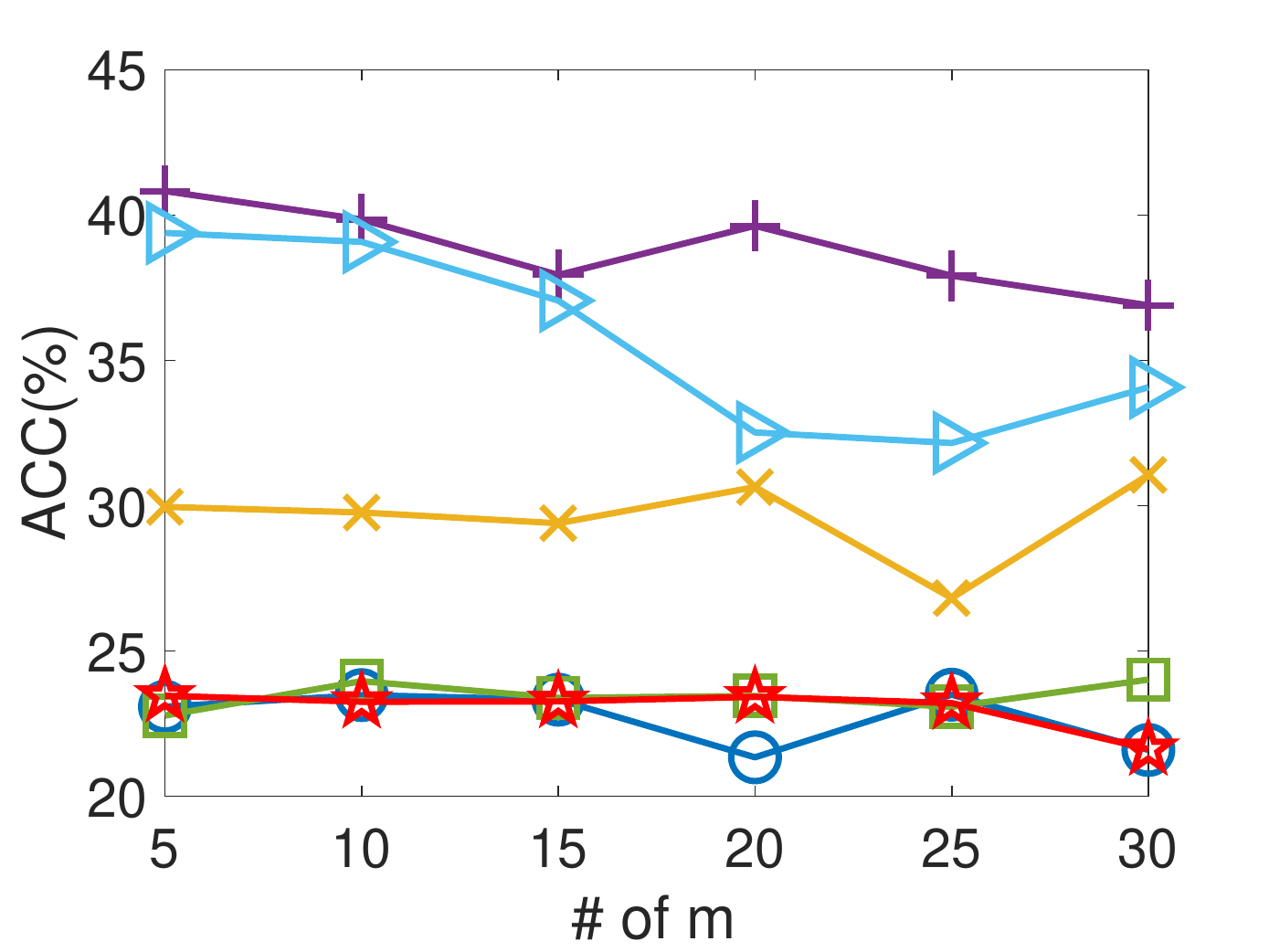}
      &\includegraphics[width=0.2\textwidth]{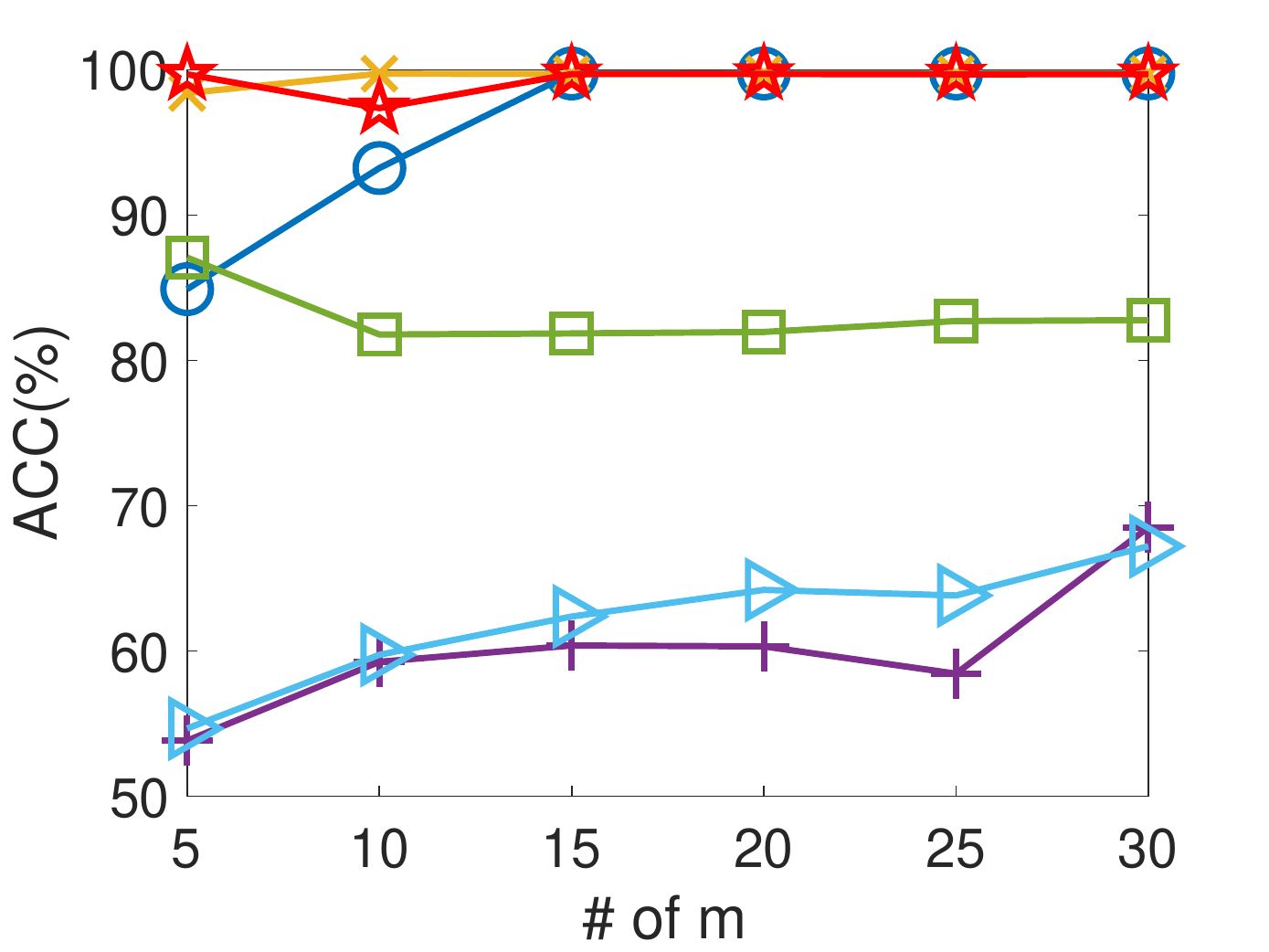}
      &\includegraphics[width=0.2\textwidth]{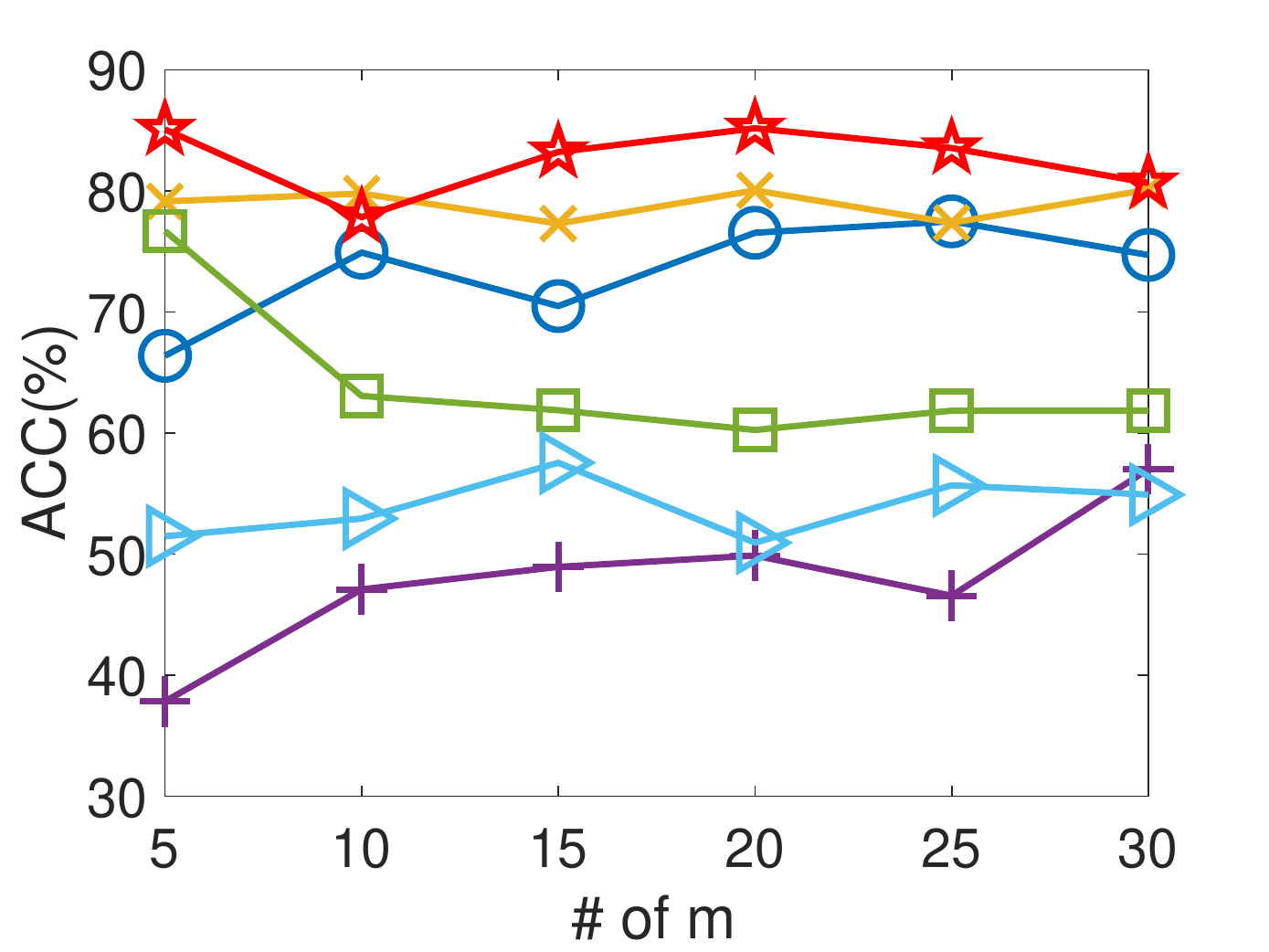}\\
      NMI
      &\includegraphics[width=0.2\textwidth]{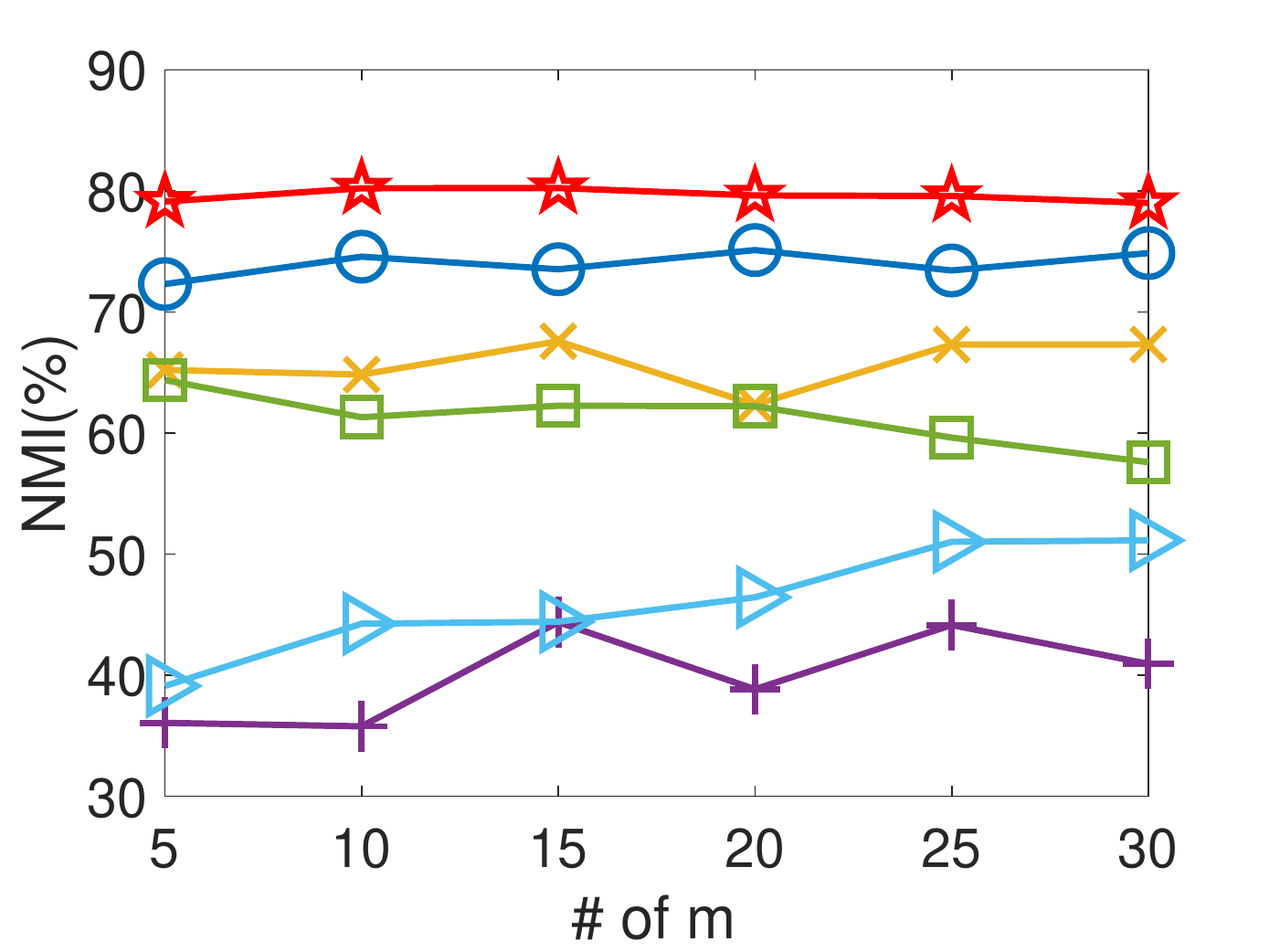}
      &\includegraphics[width=0.2\textwidth]{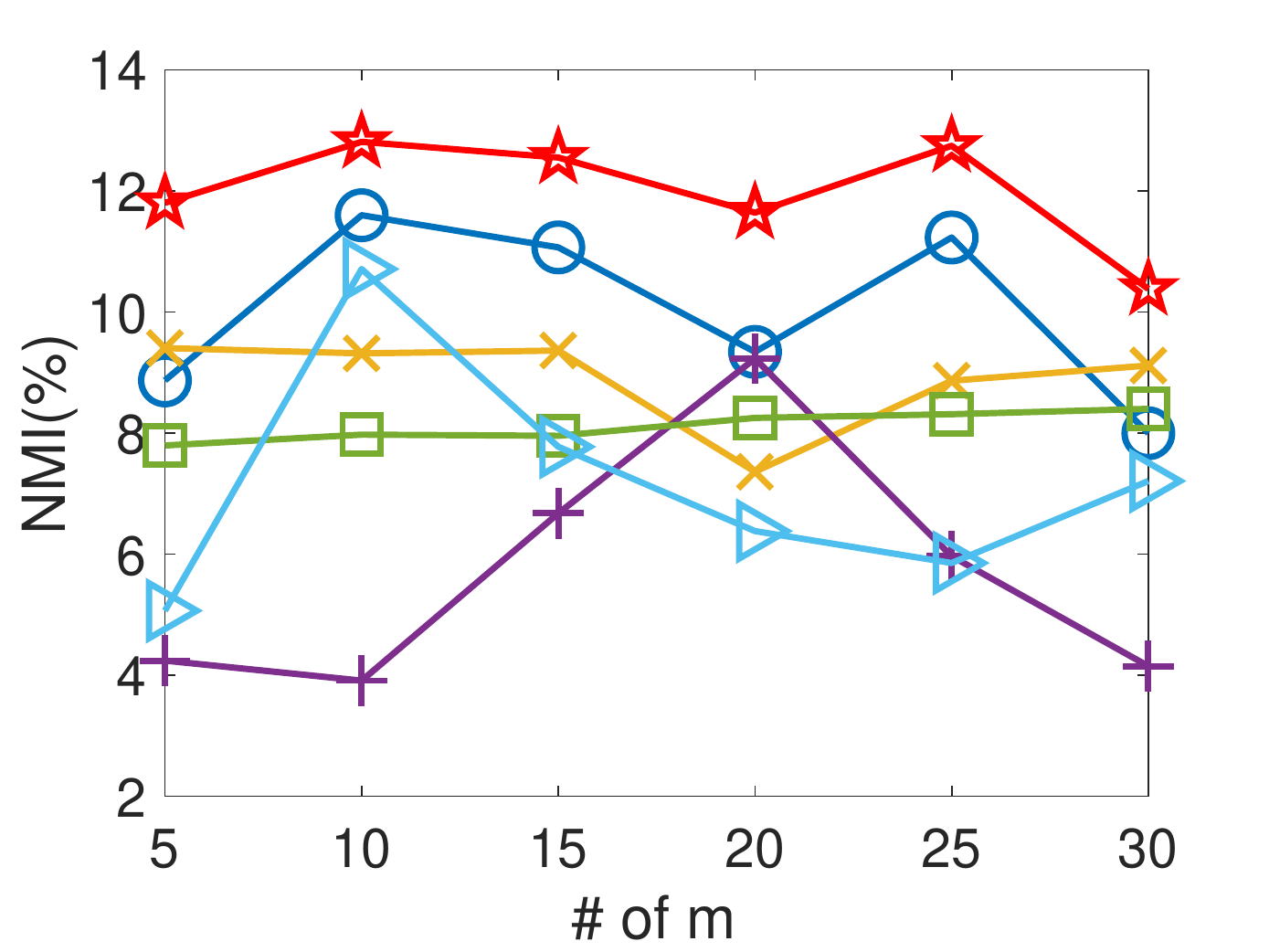}
      &\includegraphics[width=0.2\textwidth]{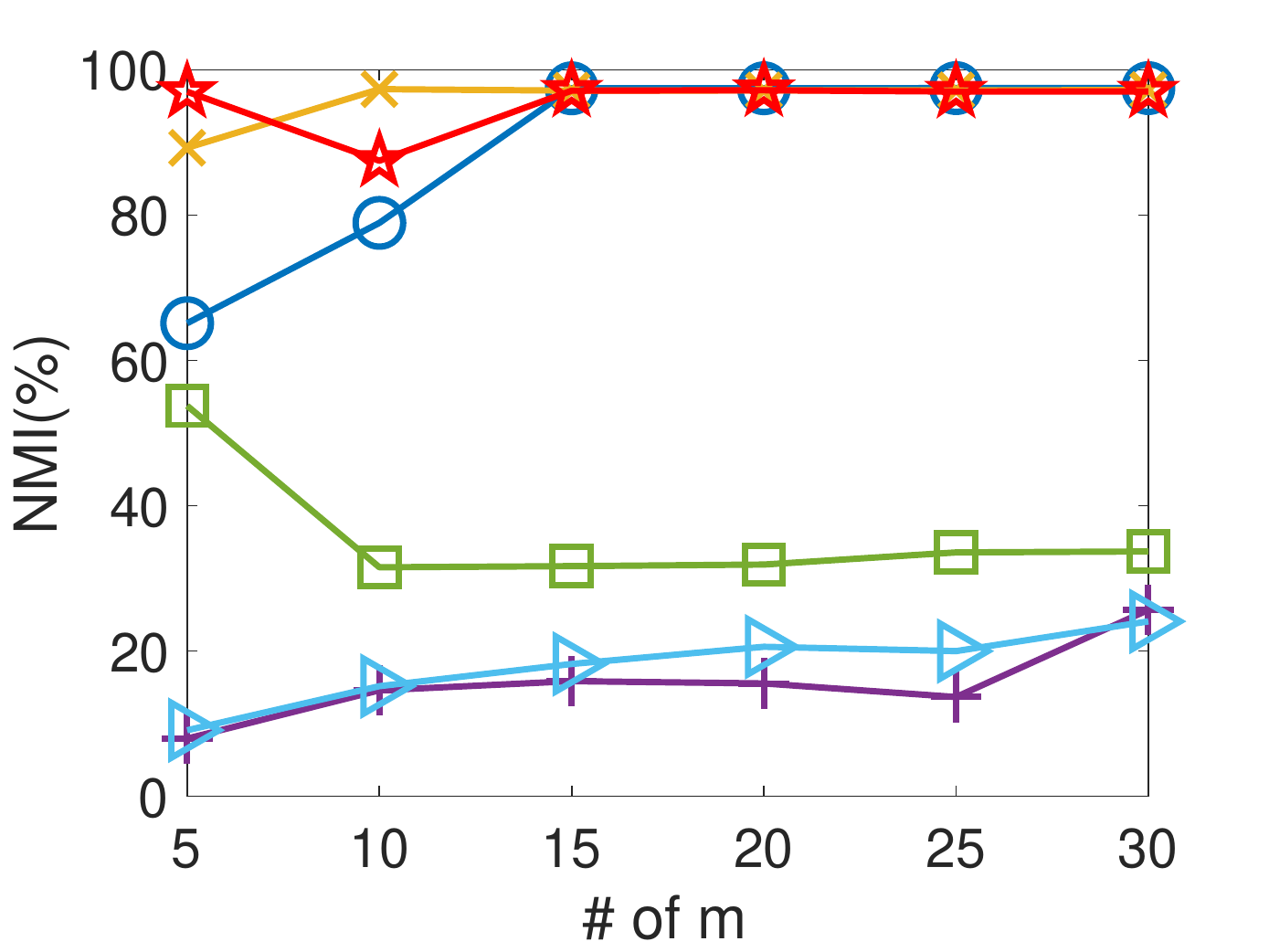}
      &\includegraphics[width=0.2\textwidth]{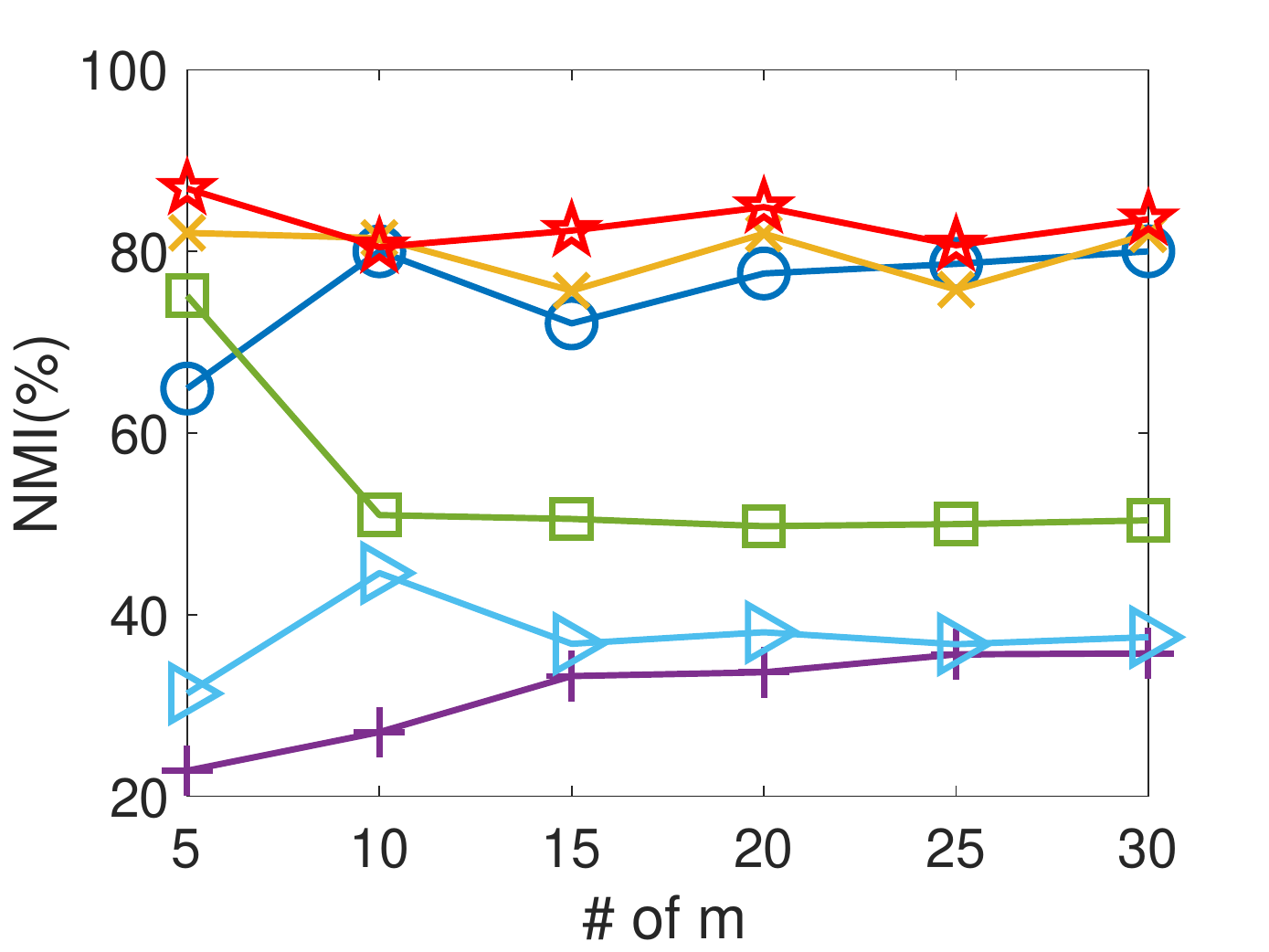}\\
      Time cost
       &\includegraphics[width=0.2\textwidth]{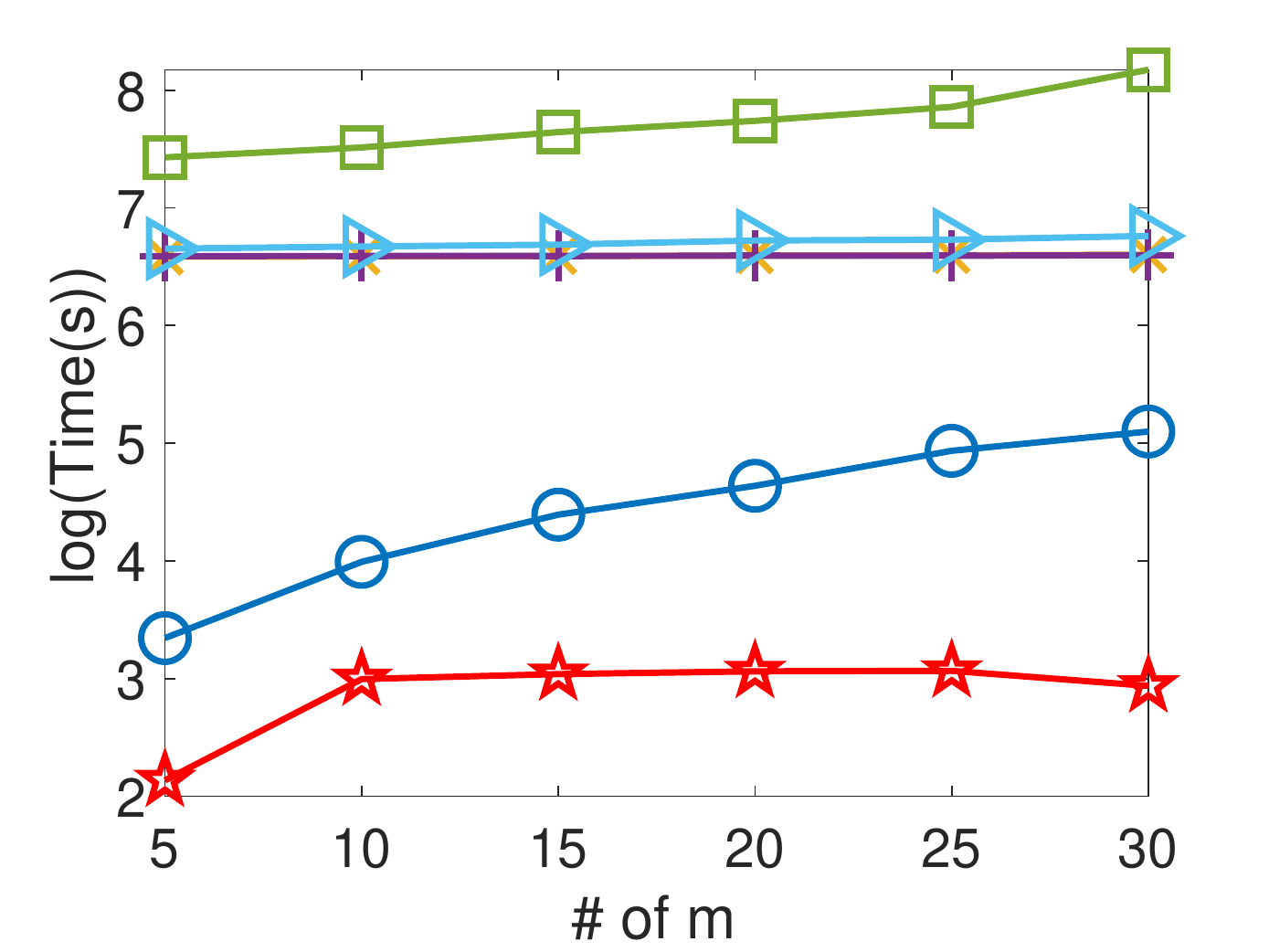}
      &\includegraphics[width=0.2\textwidth]{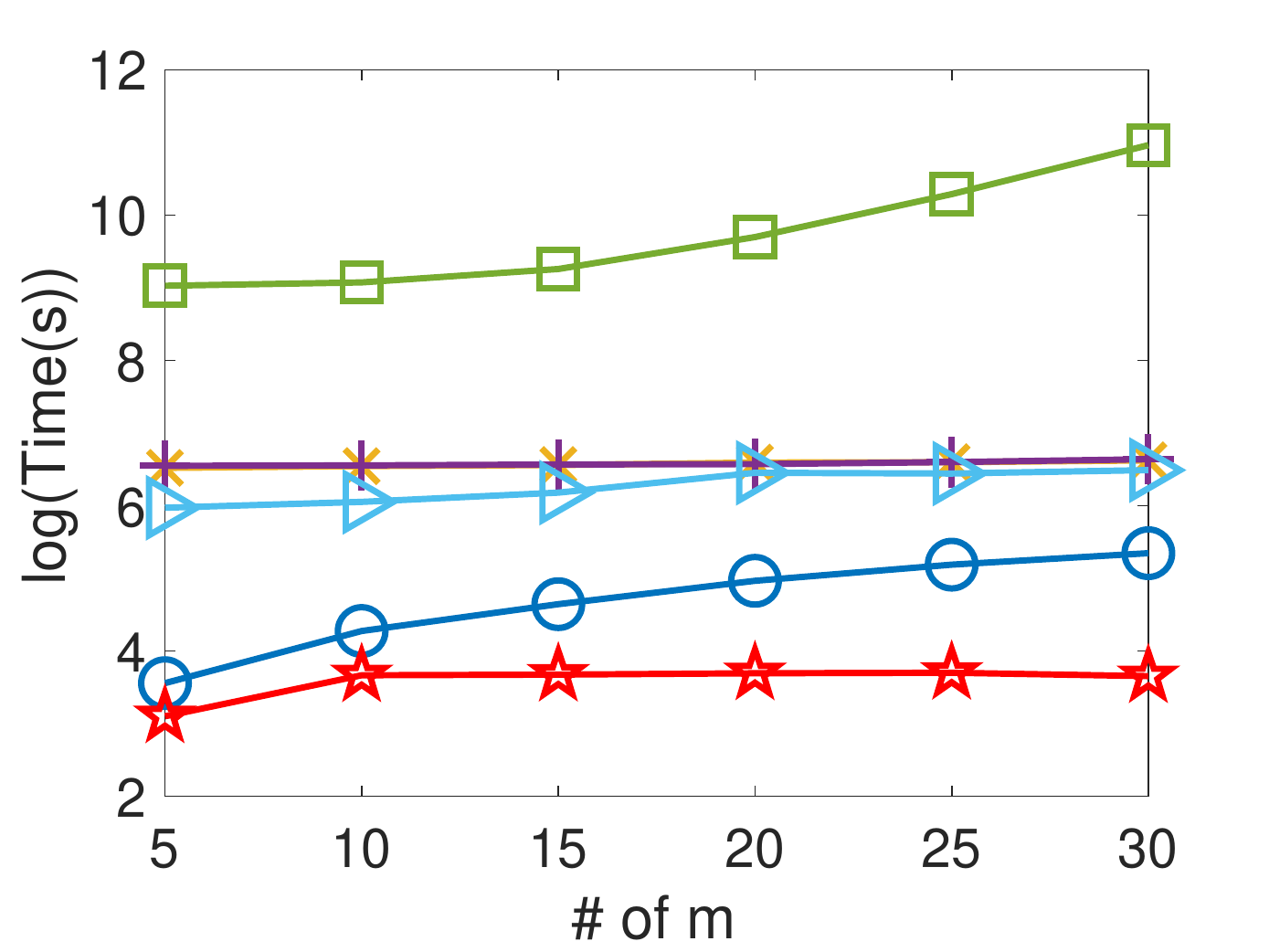}
      &\includegraphics[width=0.2\textwidth]{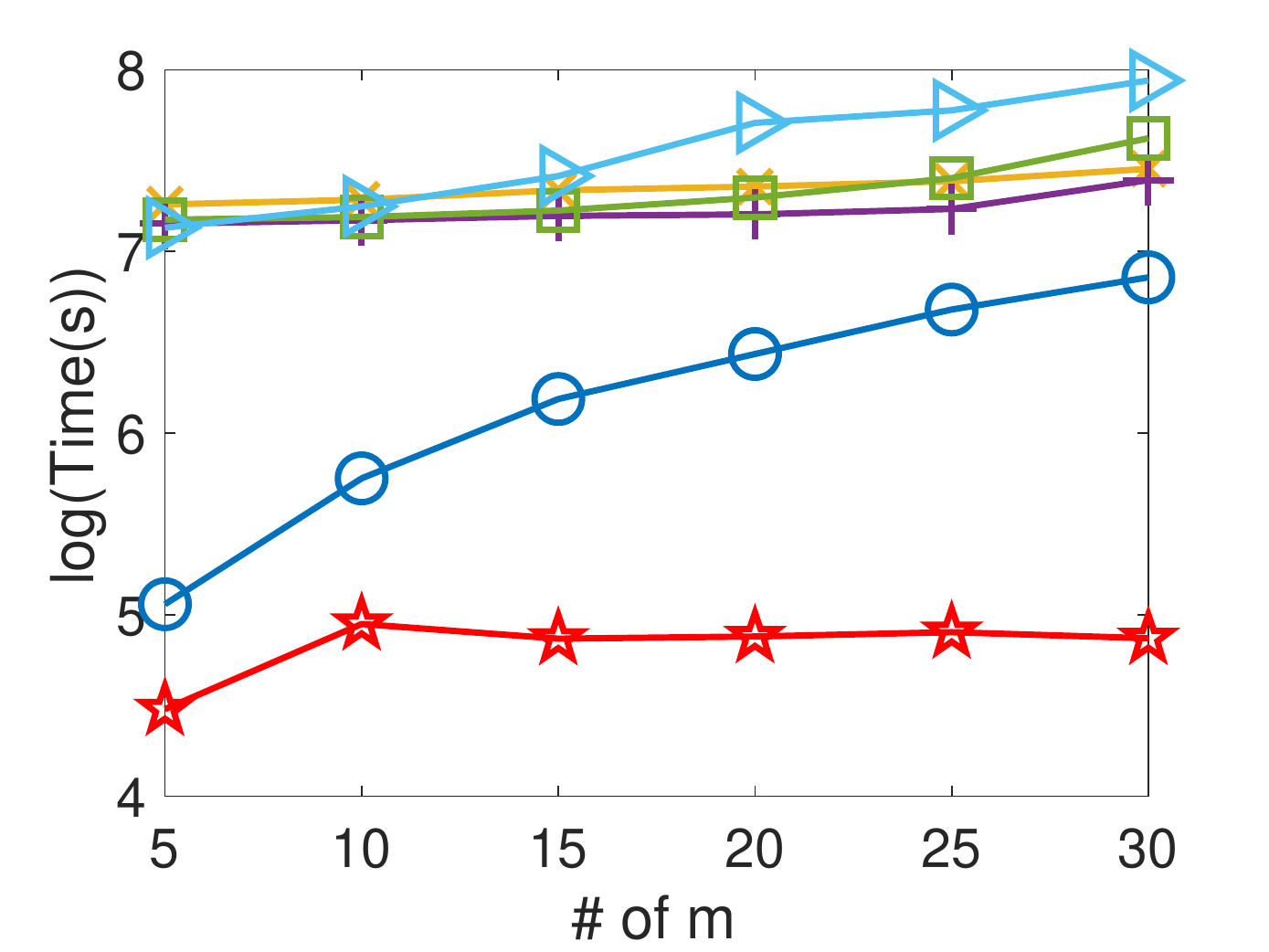}
      &\includegraphics[width=0.2\textwidth]{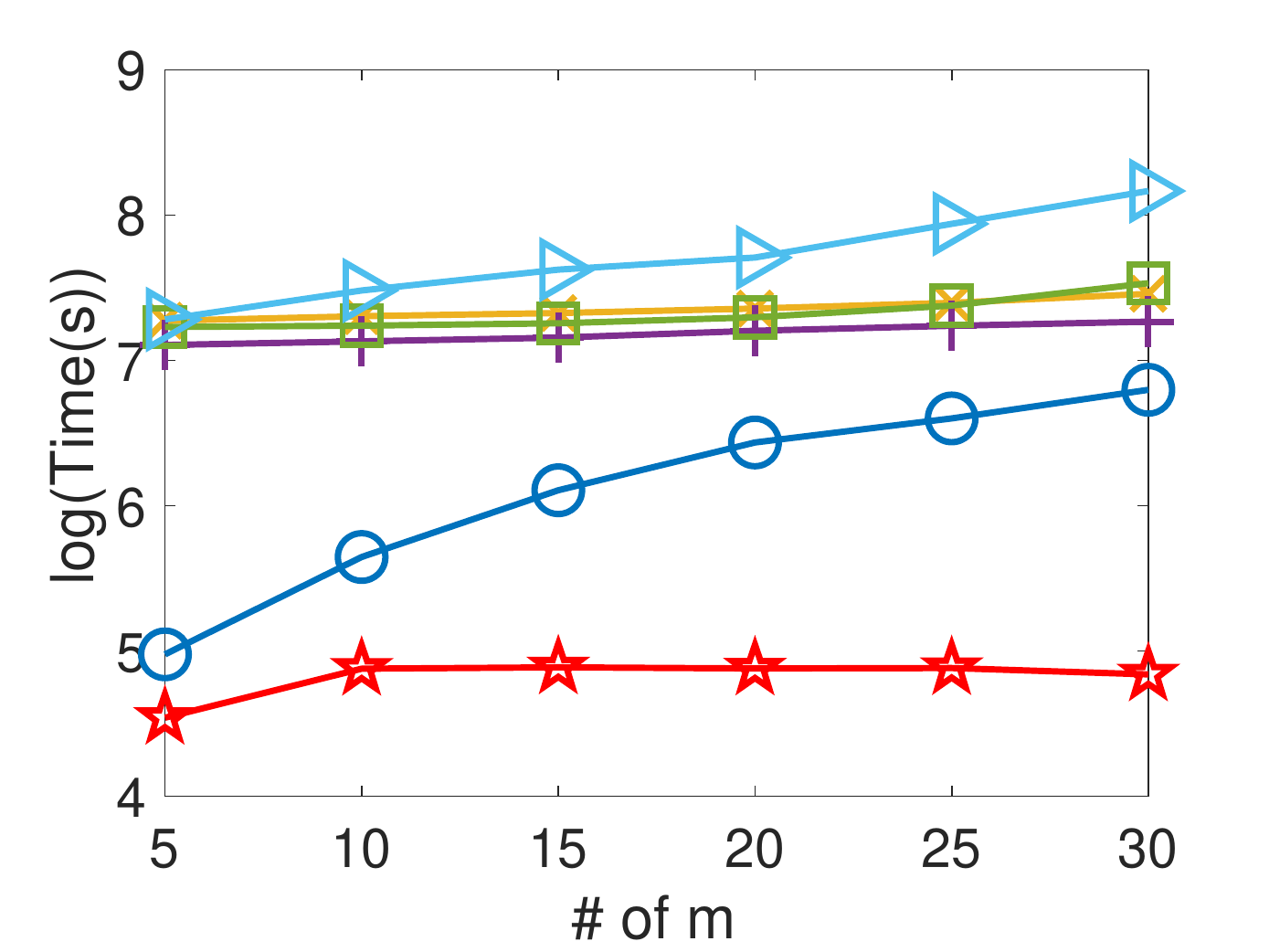}\\
       &\multicolumn{4}{c}{\includegraphics[width=0.5\textwidth]{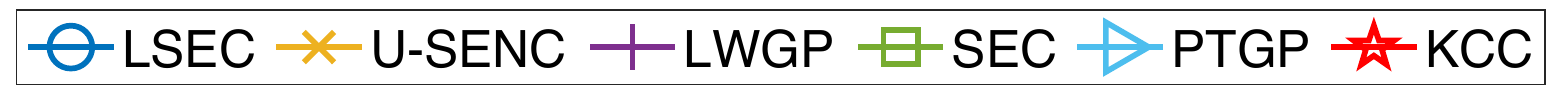}}\\
      \bottomrule
    \end{tabular}
  \end{threeparttable}
\end{table*}

\begin{table}[tpb]
  \centering
  \caption{Clustering performance (ACC(\%), NMI(\%), and time costs(s)) for LSEC with or without reusing of nearest landmarks.}
  \label{table:compare_recycle_distance}
  \begin{threeparttable}
    \begin{tabular}{m{0.75cm}<{\centering}|m{1.45cm}<{\centering}m{1.45cm}<{\centering}m{1.45cm}<{\centering}m{1.45cm}<{\centering}}
      \toprule
      \emph{Data} & \emph{MNIST} & \emph{Covertype} & \emph{TB-1M} & \emph{SF-2M} \\
      \midrule
      \multirow{1}{*}{ACC}
      &\includegraphics[width=1.7cm]{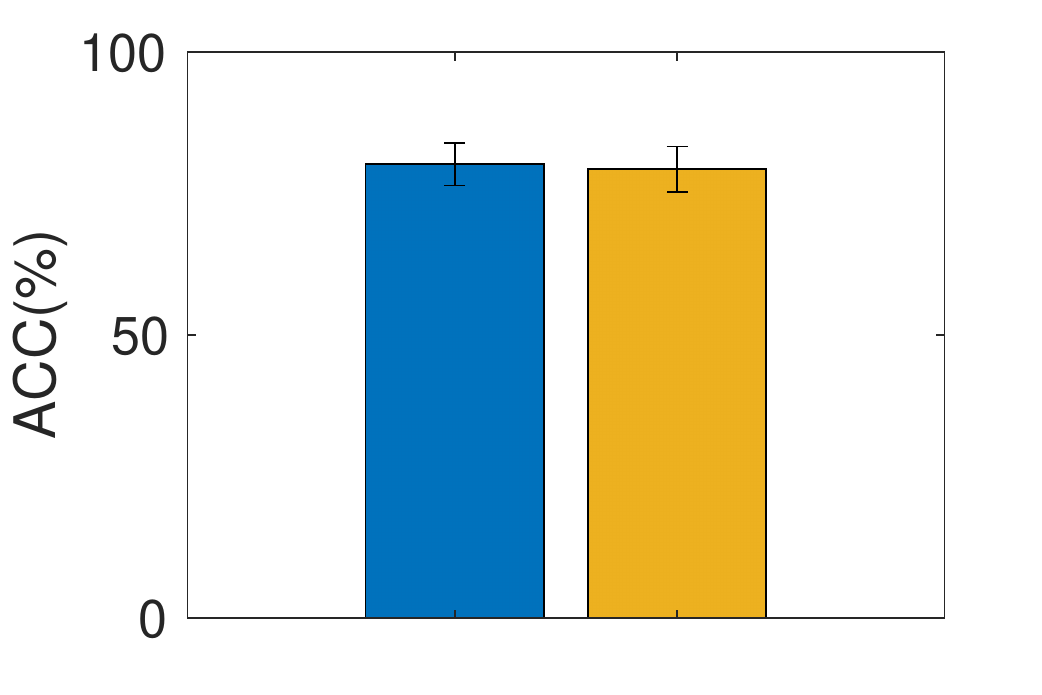}
      &\includegraphics[width=1.7cm]{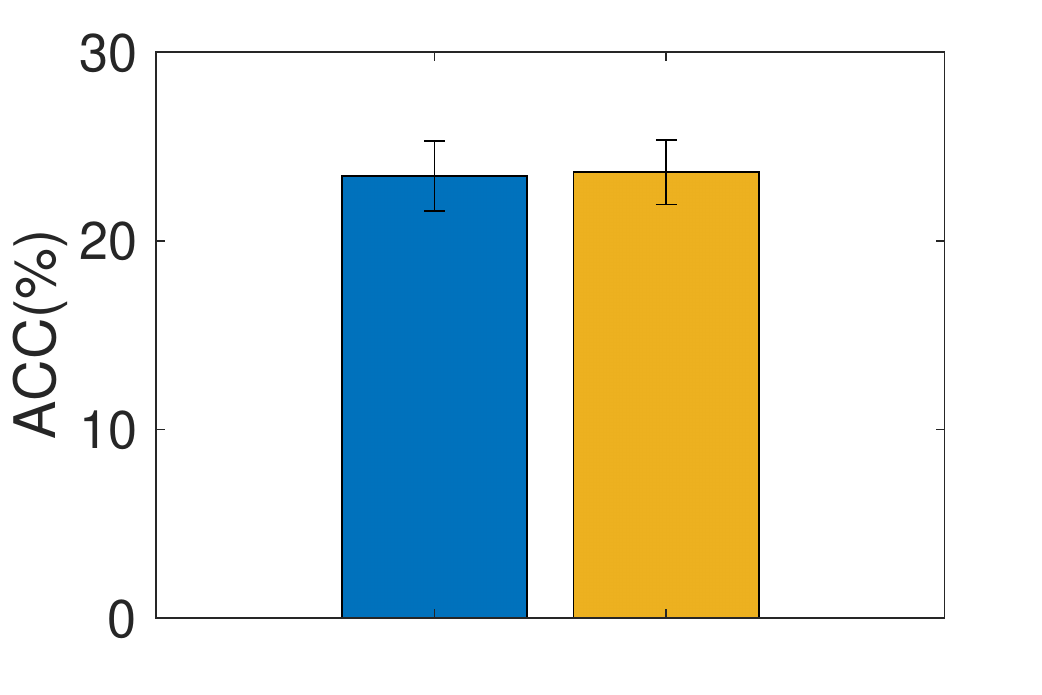}
      &\includegraphics[width=1.7cm]{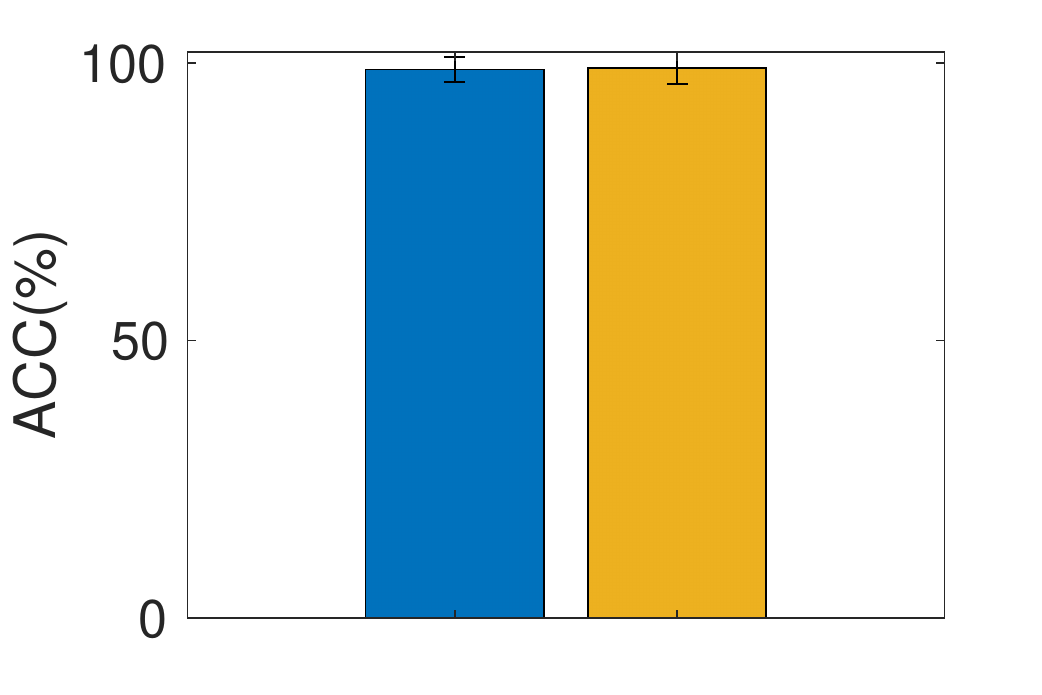}
      &\includegraphics[width=1.7cm]{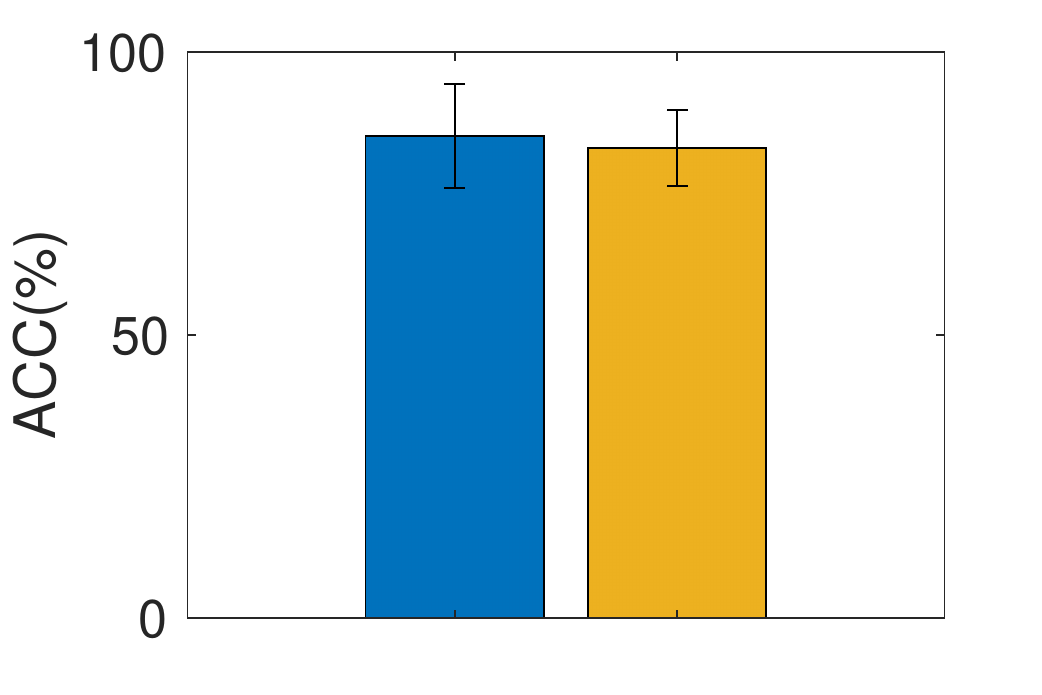}\\
      NMI
      &\includegraphics[width=1.7cm]{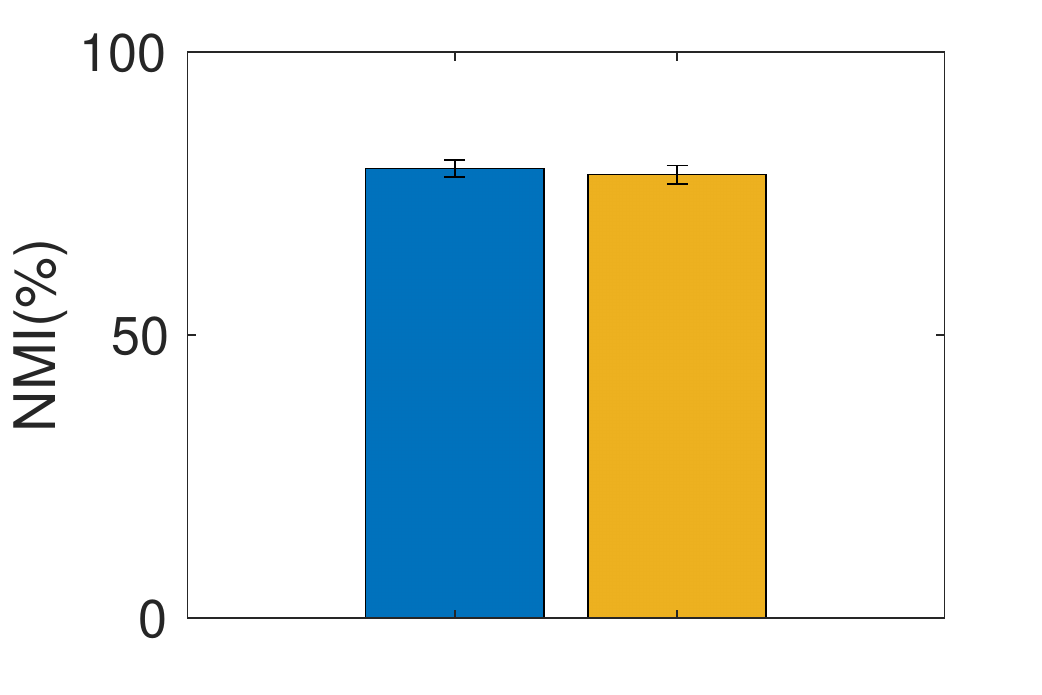}
      &\includegraphics[width=1.7cm]{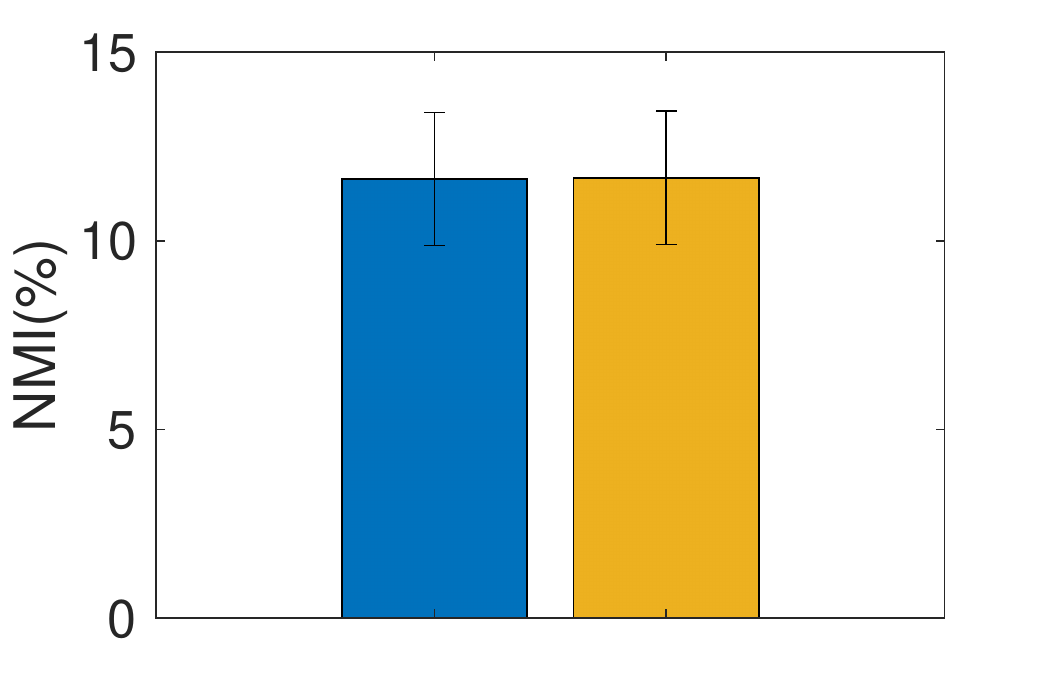}
      &\includegraphics[width=1.7cm]{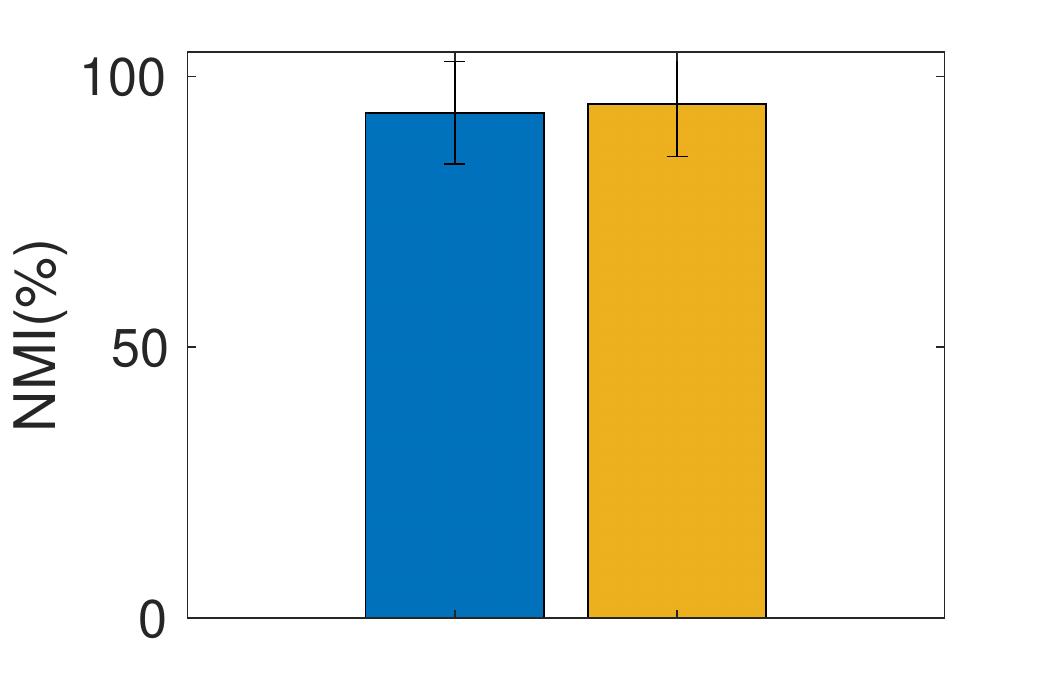}
      &\includegraphics[width=1.7cm]{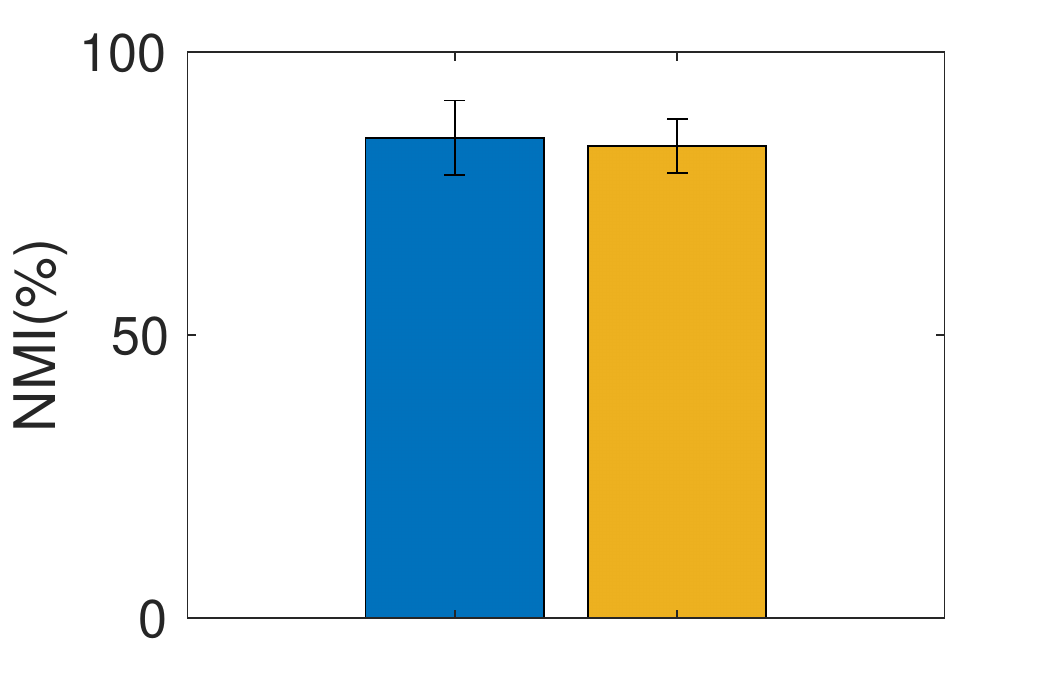}\\
      Time cost
      &\includegraphics[width=1.7cm]{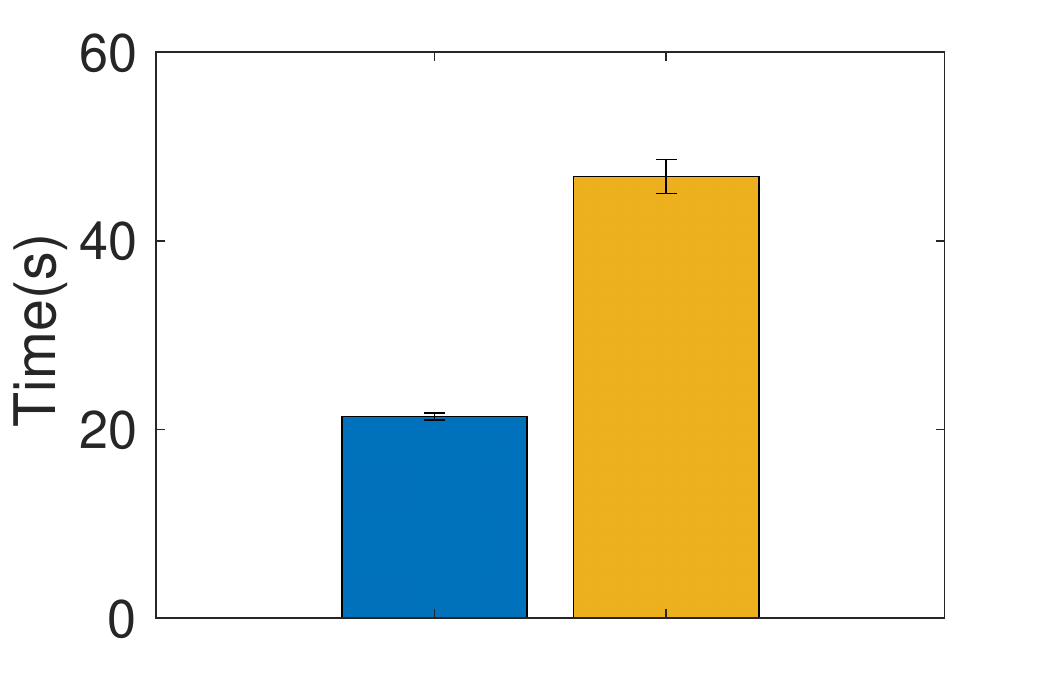}
      &\includegraphics[width=1.7cm]{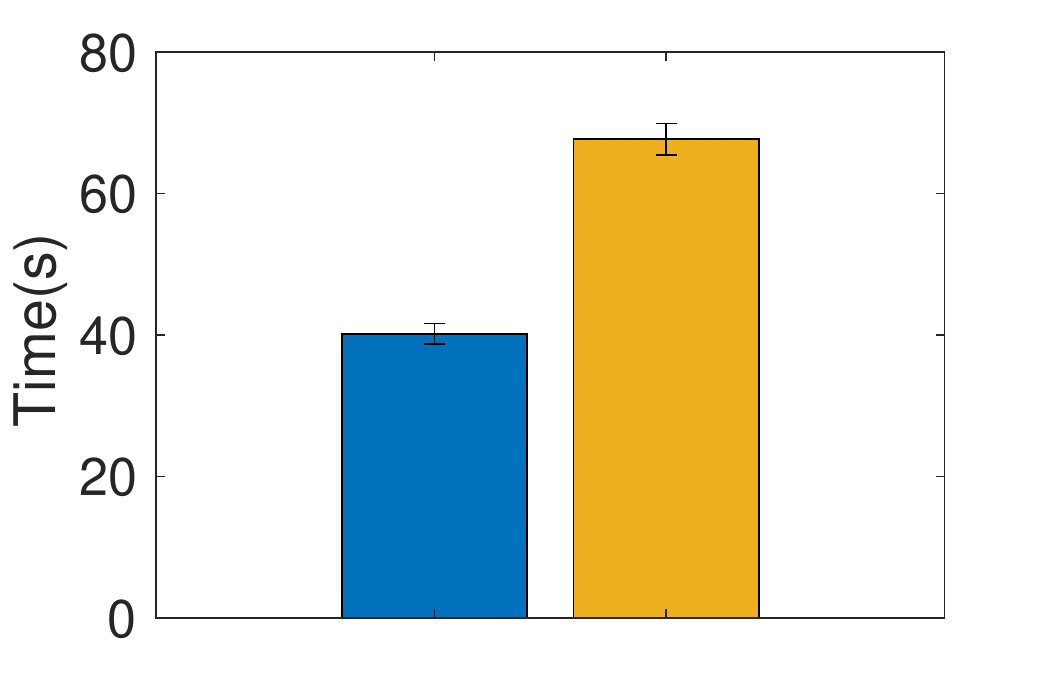}
      &\includegraphics[width=1.7cm]{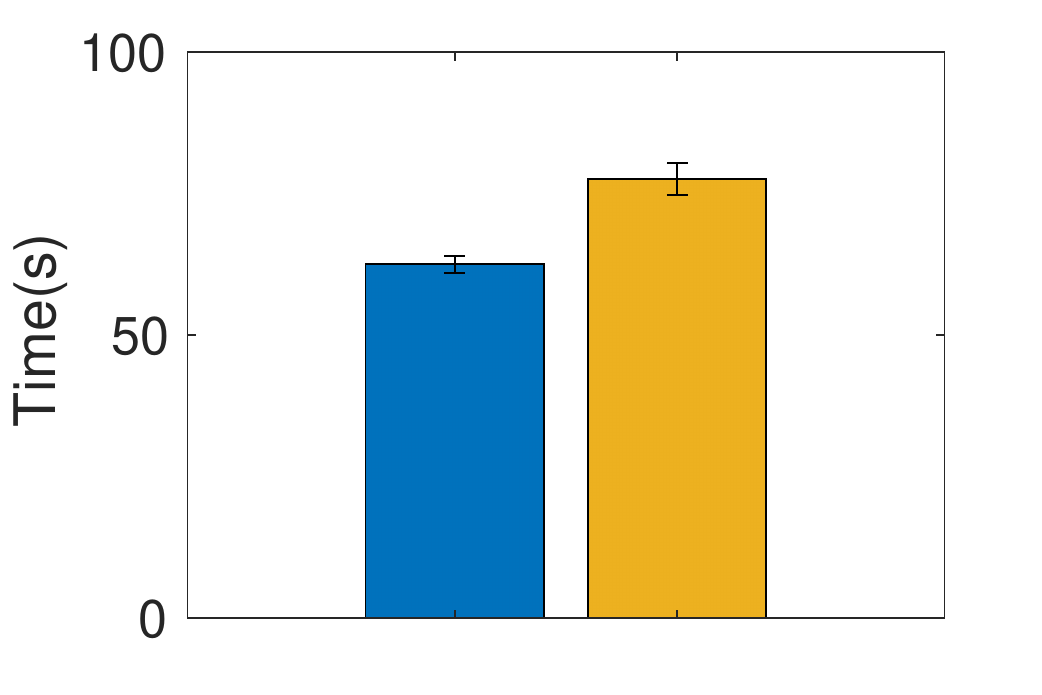}
      &\includegraphics[width=1.7cm]{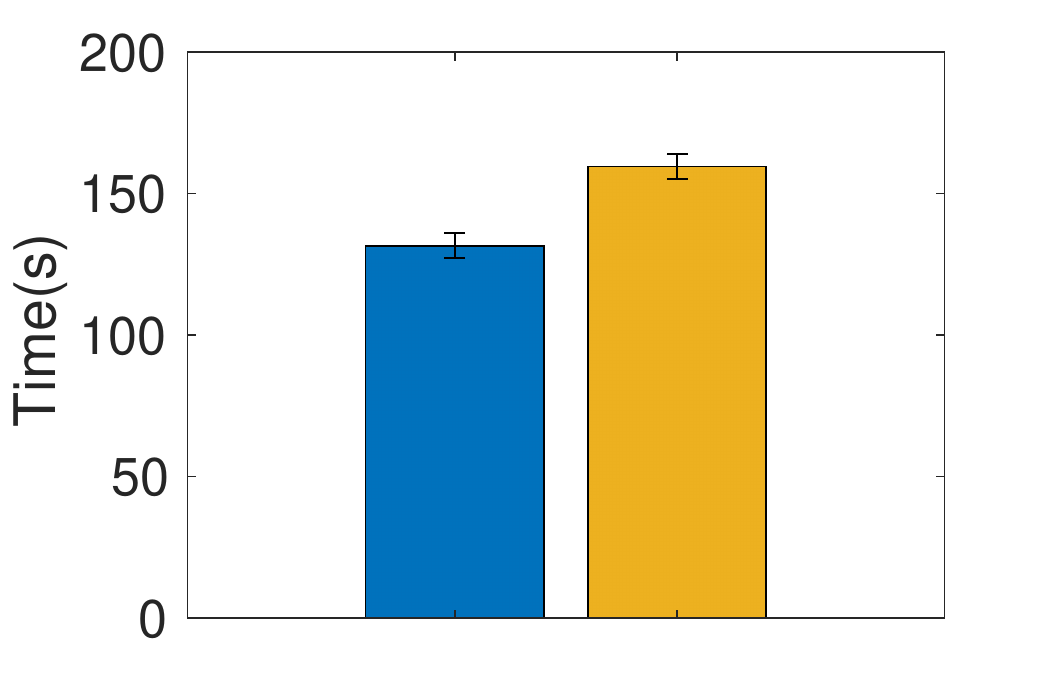}\\
      &\multicolumn{4}{c}{\includegraphics[width=0.2\textwidth]{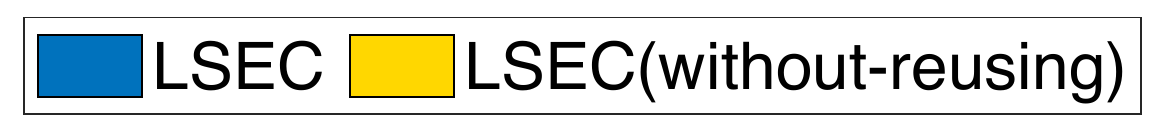}}\\
      \bottomrule
    \end{tabular}
  \end{threeparttable}
\end{table}
\begin{table}[tpb]
  \centering
  \caption{Clustering performance (ACC(\%), NMI(\%), and time costs(s)) for LSEC using light-$k$-means or using $k$-means to obtain base clusterings in the ensemble generation.}
  \label{table:compare_influnce_light_k_means}
  \begin{threeparttable}
    \begin{tabular}{m{0.75cm}<{\centering}|m{1.45cm}<{\centering}m{1.45cm}<{\centering}m{1.45cm}<{\centering}m{1.45cm}<{\centering}}
      \toprule
      \emph{Data} & \emph{MNIST} & \emph{Covertype} & \emph{TB-1M} & \emph{SF-2M} \\
      \midrule
      \multirow{1}{*}{ACC}
      &\includegraphics[width=1.7cm]{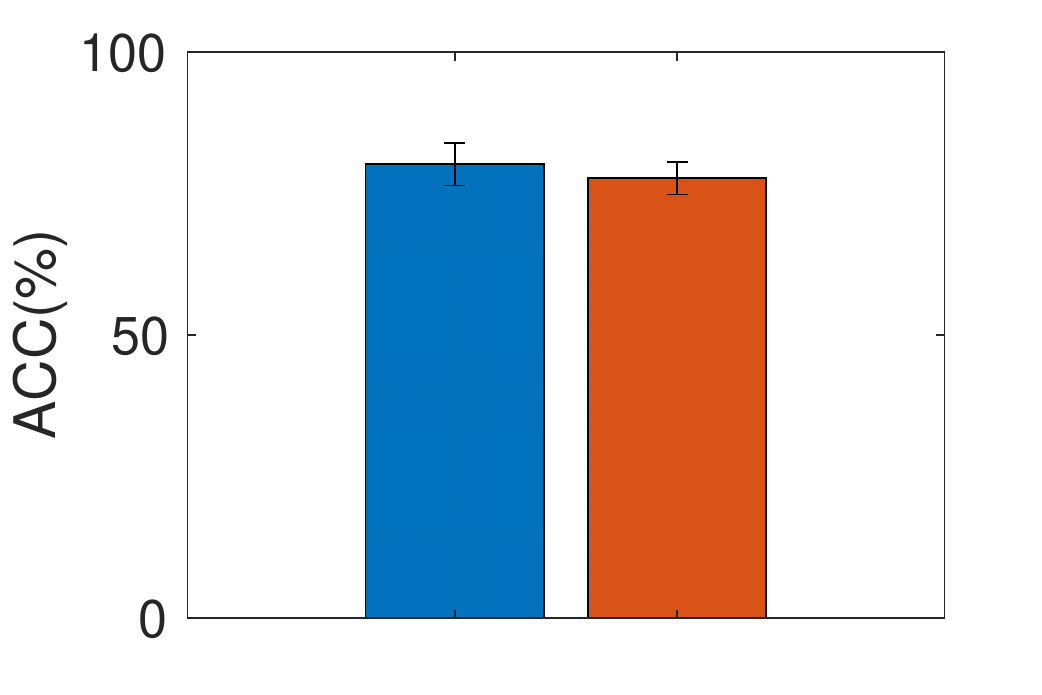}
      &\includegraphics[width=1.7cm]{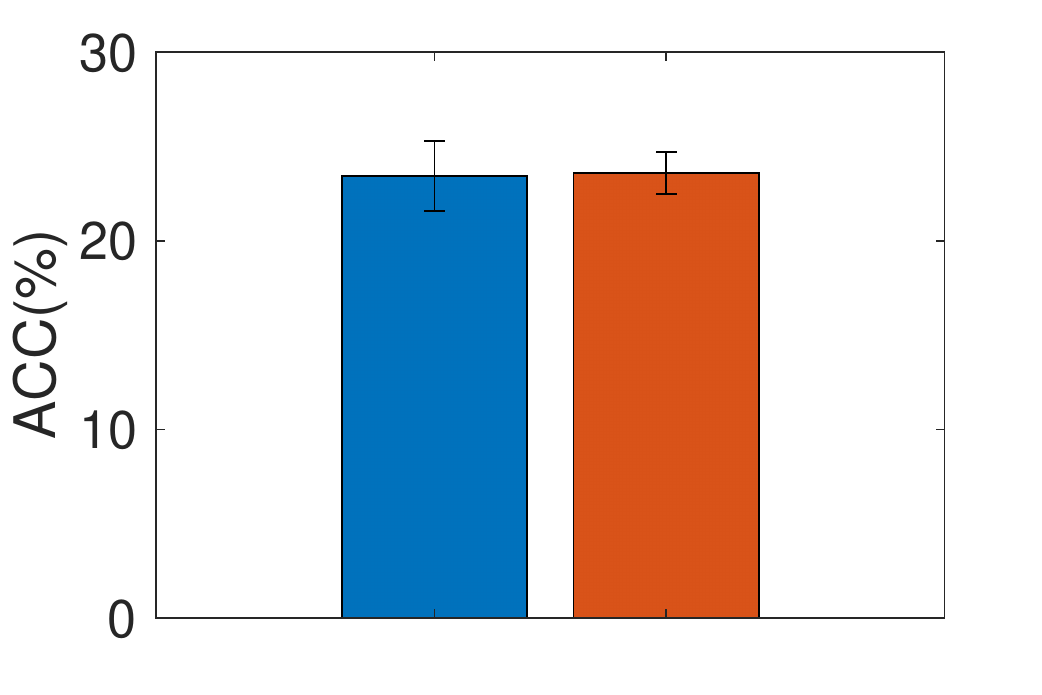}
      &\includegraphics[width=1.7cm]{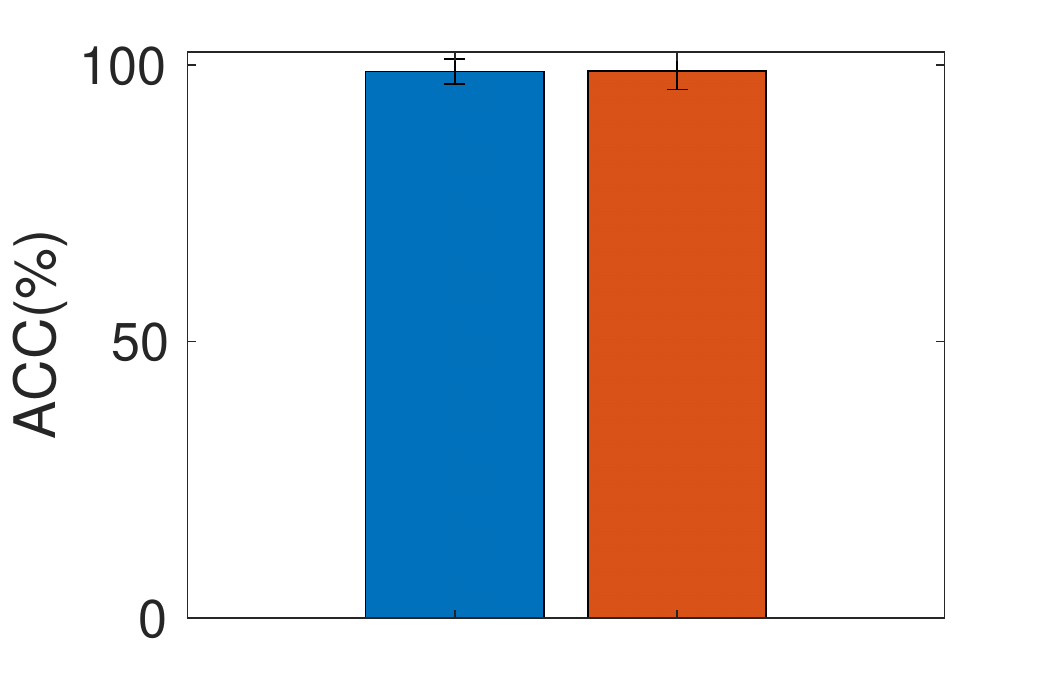}
      &\includegraphics[width=1.7cm]{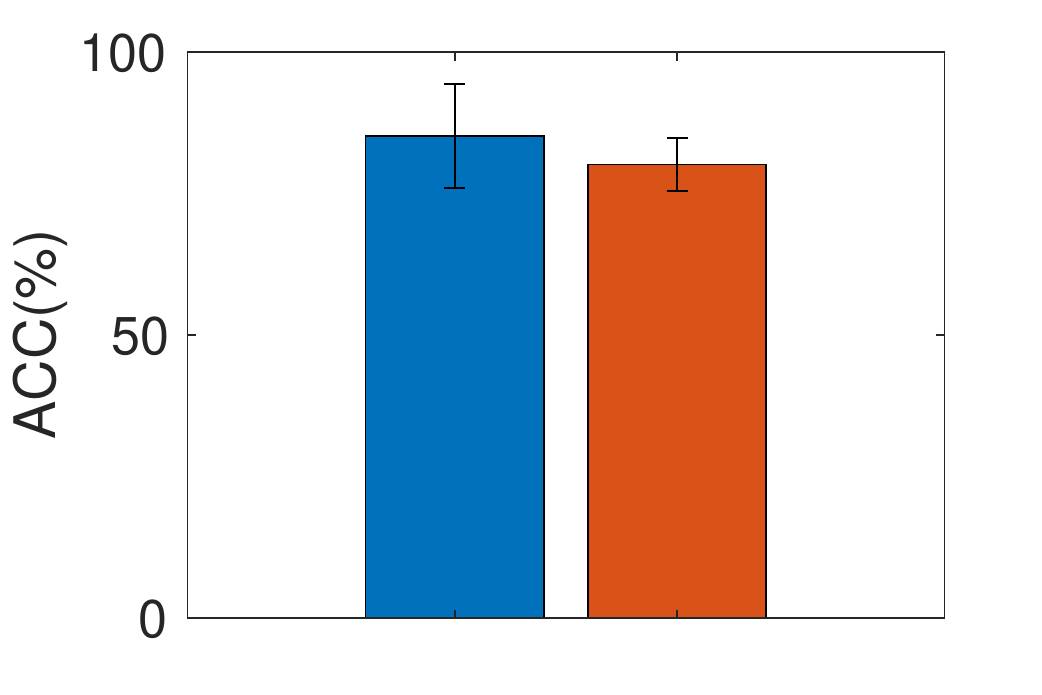}\\
      NMI
      &\includegraphics[width=1.7cm]{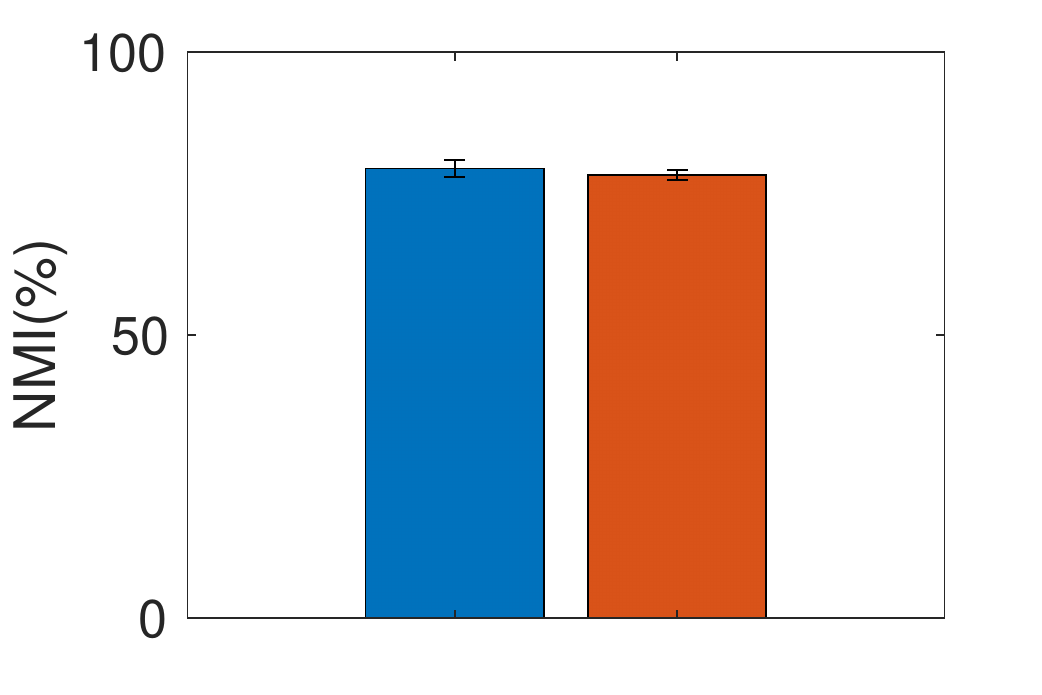}
      &\includegraphics[width=1.7cm]{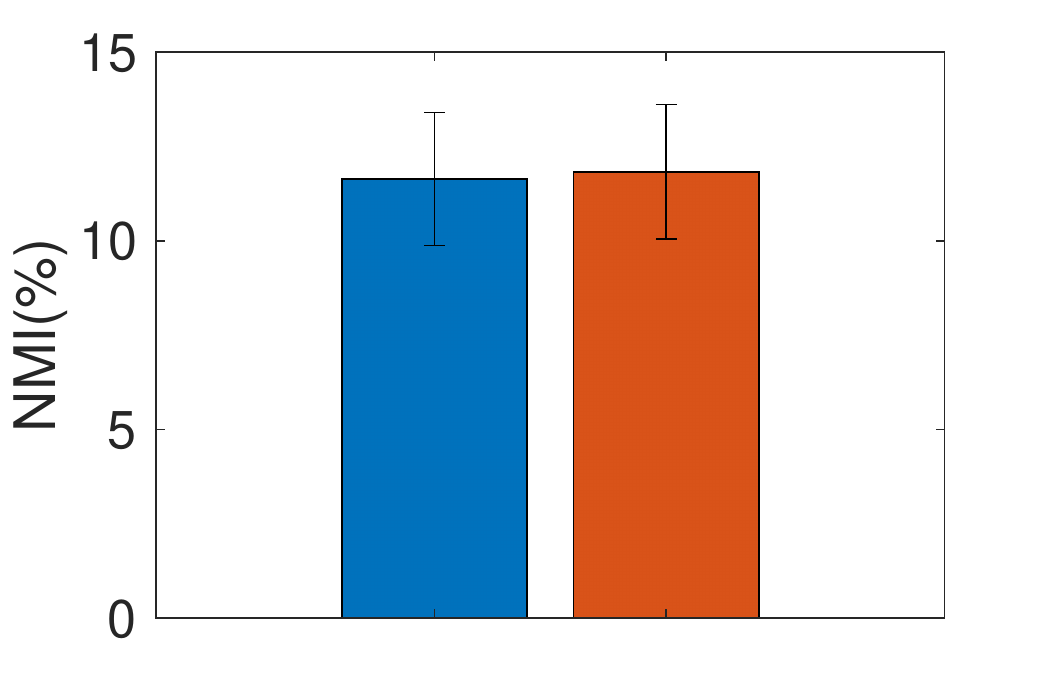}
      &\includegraphics[width=1.7cm]{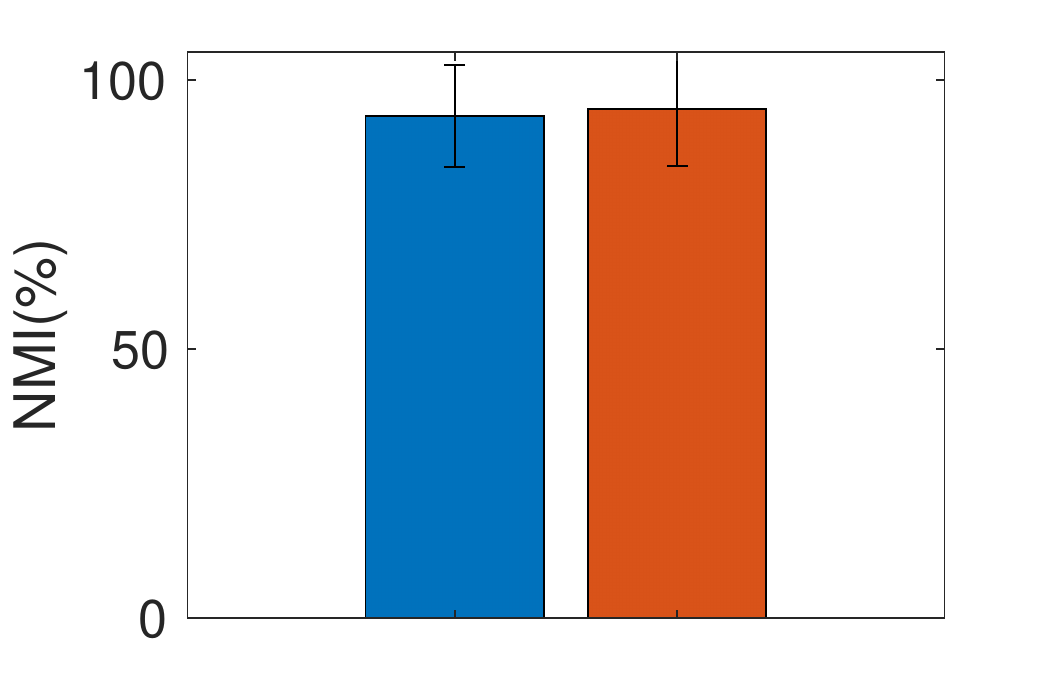}
      &\includegraphics[width=1.7cm]{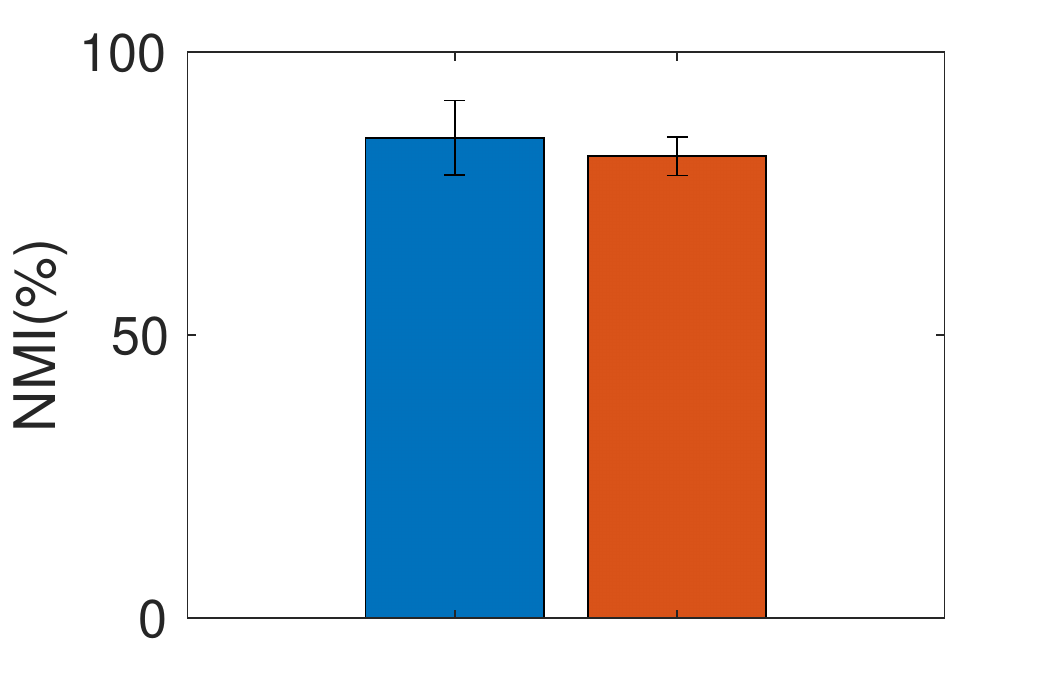}\\
      Time cost
      &\includegraphics[width=1.7cm]{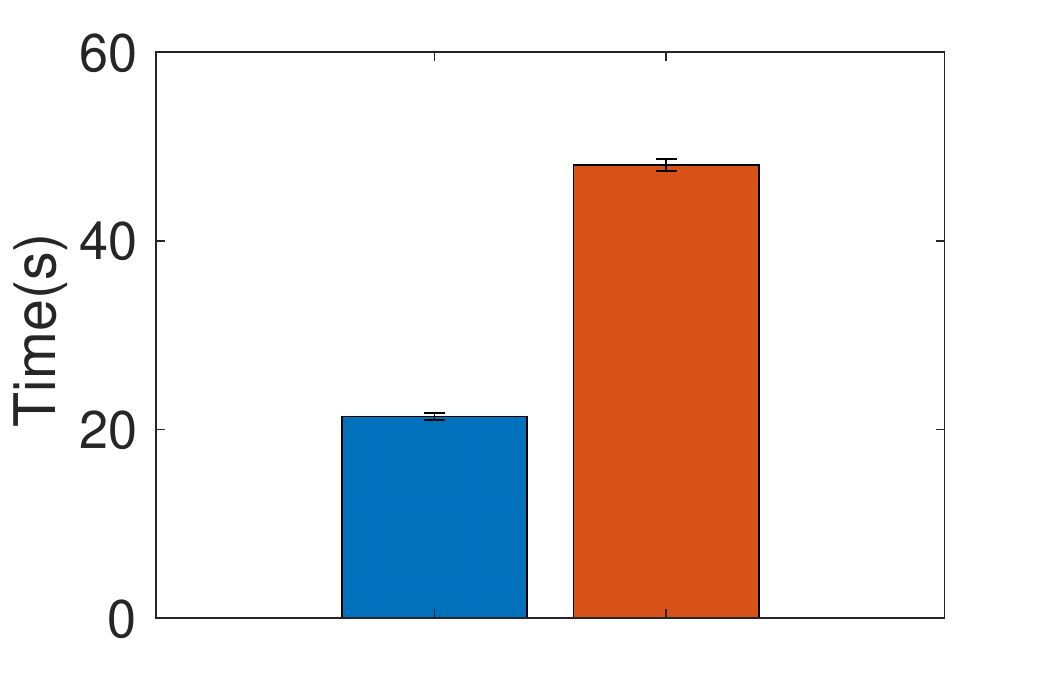}
      &\includegraphics[width=1.7cm]{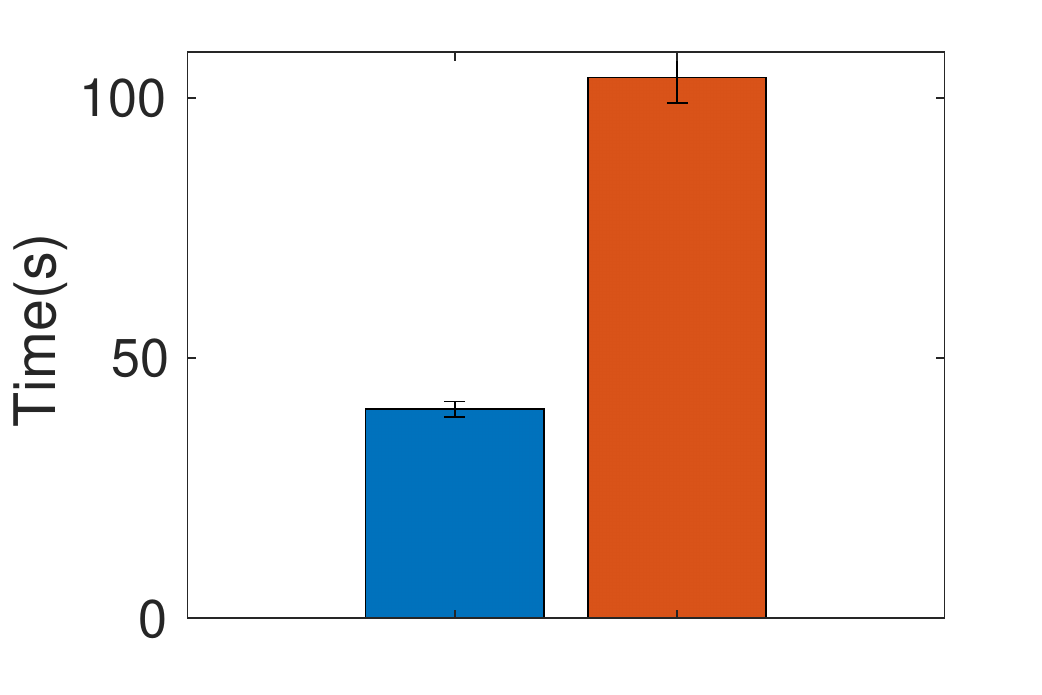}
      &\includegraphics[width=1.7cm]{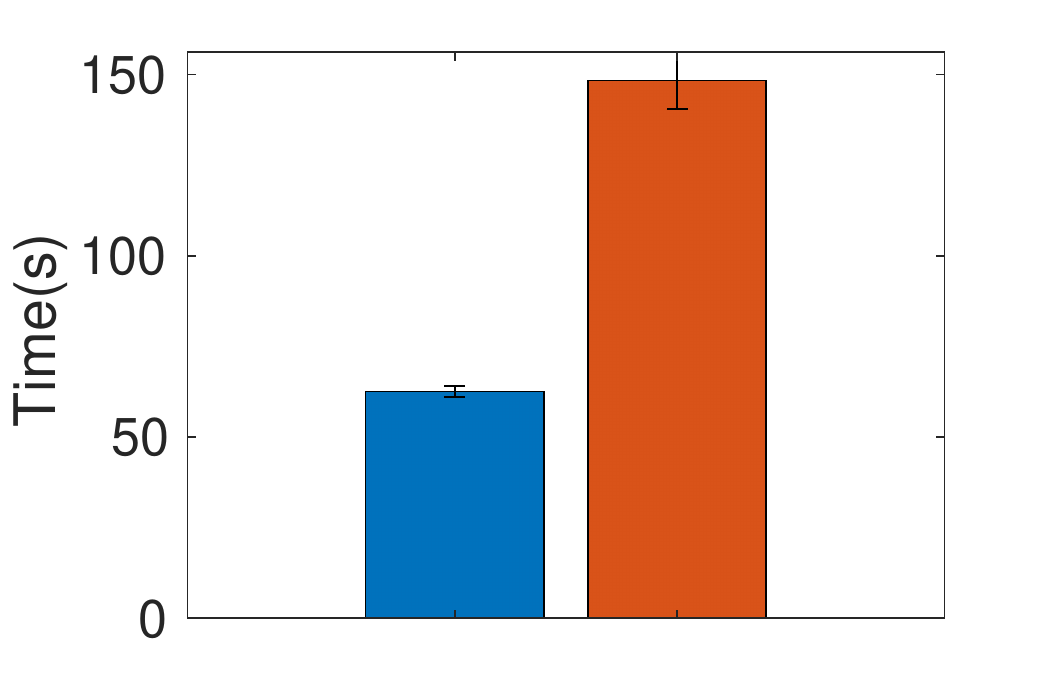}
      &\includegraphics[width=1.7cm]{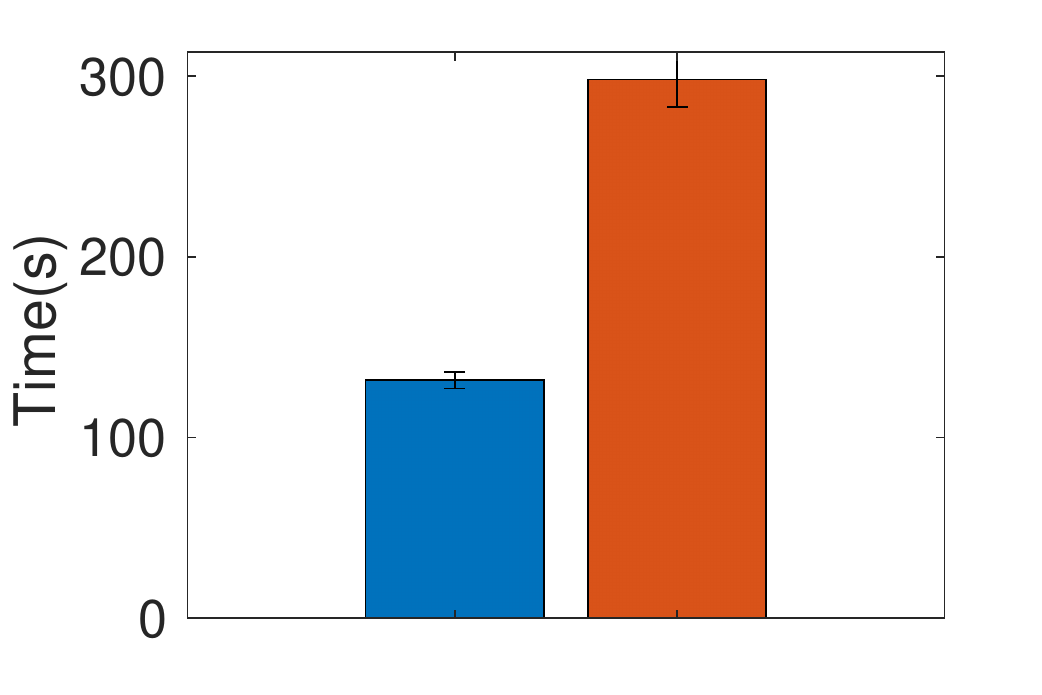}\\
      &\multicolumn{4}{c}{\includegraphics[width=0.2\textwidth]{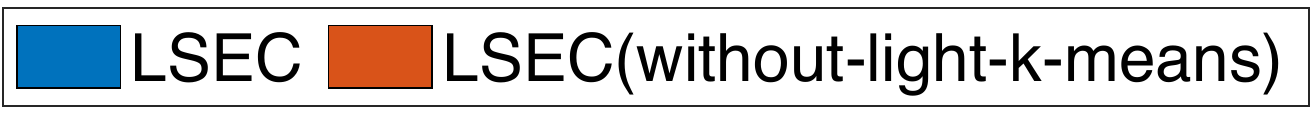}}\\
      \bottomrule
    \end{tabular}
  \end{threeparttable}
\end{table}
The experimental comparison results are reported in Tables~\ref{table:compare_ens_acc}, \ref{table:compare_ens_nmi}, and \ref{table:compare_ens_time}.
Note that DnC-SC is not an ensemble clustering algorithm; its clustering results are provided for reference only. 

As shown in Tables~\ref{table:compare_ens_acc} and \ref{table:compare_ens_nmi}, our LSEC algorithm obtains the highest ACC and NMI scores on most of datasets. In terms of average score across the ten datasets, LSEC achieves the best average ACC($\%$) and NMI($\%$) scores of $77.01$ and $77.21$, respectively.
While the second-best ensemble clustering method (i.e., U-SENC) achieves average ACC($\%$) and NMI($\%$) scores of $72.59$ and $74.78$, respectively.
The EAC, KCC, PTGP, SEC, LWGP methods use the $k$-means based ensemble generation method.
The LSEC and U-SENC methods that use the spectral clustering based ensemble generation show better clustering quality of ACC 
and NMI than others on most datasets.
In terms of average rank, LSEC obtains an average rank of 1.90 w.r.t ACC and 1.40 w.r.t. NMI, while the second-best method obtains an average rank of 2.30 w.r.t. ACC and 1.90 w.r.t. NMI.

In Table~\ref{table:compare_ens_time}, the time costs of different ensemble clustering methods are provided. 
The proposed LSEC method achieves the lowest time costs on nine datasets and the second-lowest time cost on one dataset.
Except \emph{PenDigits } dataset, LSEC is 2.4 (\emph{FL-20M}) to 5.75 (\emph{CC-5M}) times ahead of the second-best method in time consumption.
The LSEC method has shown its significant advantage over other ensemble clustering methods, especially on large-scale datasets.

\subsection{Parameter analysis on Ensemble Size $m$}
\label{sec:para_M}

We conduct a parameter analysis experiment to demonstrate the performance of the proposed method, varying different parameter values of $m$. The parameter $m$ denotes the number of base clusterings, which is a common parameter in all ensemble clustering methods.
We select four dataset (\emph{MNIST}, \emph{Covertype}, \emph{TB-1M} and \emph{SF-2M}) as benchmark datasets to conduct the following experiments. 
 As shown in Table~\ref{table:compare_para_Msize}, LSEC shows better performance of ACC and NMI than most other ensemble clustering methods except ACC score on \emph{Covertype} dataset.
Meanwhile, LSEC consistently requires a lower computational cost than all other ensemble clustering methods.

\subsection{Influence of reusing of $K$-nearest landmarks}
\label{sec:comreusing}
In this section, we compare the performances of the proposed method with or without reusing of nearest landmarks, denoted as LSEC and LSEC-without-reusing.
The experimental results are reported in Table~\ref{table:compare_recycle_distance}.
As we mentioned, the reusing of nearest landmarks brings better efficiency in searching $K$-nearest landmarks.
In Table~\ref{table:compare_recycle_distance}, LSEC and LSEC-without-reusing show similar performances to each other, but LSEC cost obviously less time.
Since reusing of nearest landmarks does not influence the accuracy of nearest landmarks, we consider that the difference of ACC and NMI between the two methods comes from the randomness of the algorithm.
This result indicates that reusing of nearest landmarks achieves significantly better efficiency while maintaining a similar clustering result.

\subsection{Influence of light-$k$-means}
\label{sec:cmplight-kmeans}
In this section, we compare the performances of the proposed method using light-$k$-means or using $k$-means to obtain base clusterings in the ensemble generation.
The experimental results are reported in Table \ref{table:compare_influnce_light_k_means}.
Generally, two methods show the similar performance of ACC and NMI.
Especially, LSEC achieves slightly better ACC and NMI on \emph{MNIST} and \emph{SF-2F} datasets, which is possible because the light-$k$-means method can provide better diversity of base clusterings on these datasets. 
Overall, using light-$k$-means in ensemble generation significantly improves the efficiency of LSEC and yields similar clustering quality compared to the $k$-means method.

\section{Conclusion}
\label{sec:conclusion}

In this paper, we propose a large-scale spectral ensemble clustering (LSEC) method to strike a better balance between the efficiency and effectiveness of ensemble clustering on large-scale datasets.
We design an efficient ensemble generation framework to produce based clustering, applying divide-and-conquer large-scale spectral clustering to find high-quality base clusterings. 
In the ensemble generation of the proposed method, we accelerate the process of searching $K$-nearest neighbors by reusing strategy and obtaining base clustering by the light-$k$-means method.
After the ensemble generation step, we combine all based clustering into a consensus cluster through a bipartite graph partitioning based consensus function.
The proposed method achieves lower computational complexity than most existing ensemble clustering methods.
Experiments conducted on ten large-scale datasets show that the proposed method outperforms other state-of-the-art large-scale spectral clustering methods.

\section*{Acknowledgment}
This study was supported by the New Energy and Industrial Technology Development Organization (NEDO) Grant (ID:18065620) and JST COI-NEXT.

\bibliographystyle{plain}
\bibliography{main}

\end{document}